%% file: iclr2026_conference.tex
\documentclass{article} 
\usepackage{iclr2026_conference, times}

\input{math_commands.tex}

\usepackage{hyperref}
\usepackage[utf8]{inputenc} 
\usepackage[T1]{fontenc}    
\usepackage{url}            
\usepackage{booktabs}       
\usepackage{multirow}       
\usepackage{nicefrac}       
\usepackage{microtype}      
\usepackage[dvipsnames,svgnames,table]{xcolor}
\usepackage{siunitx}  
\usepackage{array}    
\usepackage{algpseudocode}
\usepackage{multicol}
\usepackage{graphicx} 
\usepackage{amssymb}
\usepackage{algorithm}
\usepackage{wrapfig}
\usepackage{subcaption}
\usepackage{amsfonts, amsmath, bbm}
\usepackage[capitalize]{cleveref}
\usepackage{makecell}
\usepackage{float} 
\usepackage{enumitem}
\usepackage{tabularx}
\usepackage{caption}
\usepackage{wrapfig}
\usepackage{hyperref}
\usepackage{soul}
\definecolor{jointAttnColor}{HTML}{1470CA}
\definecolor{moeColor}{HTML}{FF1F0B}      
\definecolor{projColor}{HTML}{2E8B57}     

\soulregister\cite7
\soulregister\citep7
\soulregister\citet7
\soulregister\ref7
\soulregister\pageref7
\soulregister\cref7
\soulregister\Cref7

\title{TwinVLA: Data-Efficient Bimanual Manipulation with Twin Single-Arm Vision-Language-Action Models}

\author{Hokyun Im\textsuperscript{1,2} \quad Euijin Jeong\textsuperscript{1} \quad Andrey Kolobov\textsuperscript{2} \quad Jianlong Fu\textsuperscript{2} \quad Youngwoon Lee\textsuperscript{1} \\
\textsuperscript{1}Department of Artificial Intelligence, Yonsei University \quad \textsuperscript{2}Microsoft Research
\\
\url{https://jellyho.github.io/TwinVLA/}
}

\iclrfinalcopy 
\begin{document}

\maketitle

\begin{abstract}
Vision-language-action models (VLAs) trained on large-scale robotic datasets have demonstrated strong performance on manipulation tasks, including bimanual tasks. However, because most public datasets focus on single-arm demonstrations, adapting VLAs for bimanual tasks typically requires substantial additional bimanual data and fine-tuning. To address this challenge, we introduce TwinVLA, a modular framework that composes two copies of a pretrained single-arm VLA into a coordinated bimanual VLA. Unlike monolithic cross-embodiment models trained on mixtures of single-arm and bimanual data, TwinVLA improves both data efficiency and performance by composing pretrained single-arm policies. Across diverse bimanual tasks in real-world and simulation settings, TwinVLA outperforms a comparably-sized monolithic RDT-1B model without requiring \emph{any} bimanual pretraining. Furthermore, it narrows the gap to state-of-the-art model $\pi_0$, which relies on extensive proprietary bimanual data and compute cost. These results establish our modular composition approach as a data-efficient and scalable path toward high-performance bimanual manipulation, leveraging public single-arm data.
\end{abstract}

\section{Introduction}

Thanks to publicly available large-scale robotic datasets, vision-language-action models (VLAs) have shown impressive performance in single-arm robotic manipulation, effectively adapting to downstream tasks and generalizing across diverse tasks, objects, and environments~\citep{rt22023arxiv, oxe2024, kim2024openvla, black2024pi_0}. However, extending these successes to \textit{bimanual} manipulation remains challenging, as public bimanual datasets are scarce, and existing approaches often rely on large, proprietary datasets that require thousands of hours of data collection and curation~\citep{black2024pi_0}, limiting reproducibility and progress.

Can we build strong bimanual VLAs without collecting or fine-tuning on large bimanual datasets by leveraging existing single-arm data? In this work, we propose a highly data-efficient adaptation paradigm for bimanual control that eliminates the need for prohibitive bimanual pretraining. By effectively repurposing a single-arm VLA, we demonstrate that complex bimanual skills can be mastered using only minimal target-domain demonstrations, establishing a practical and reproducible pathway to bimanual manipulation.

To effectively realize this transfer of single-arm priors to bimanual control, the choice of underlying architecture is critical. Recent cross-embodiment learning work typically trains monolithic models on multi-robot datasets~\citep{oxe2024} employing embodiment-specific action decoders~\citep{octo_2023, nvidia2025gr00tn1openfoundation} or shared, zero-padded action spaces~\citep{liu2024rdt, black2024pi_0}. Although promising, differences in observation and action spaces introduce heterogeneity, forcing a single model to handle disparate action spaces, and monolithic training underutilizes the \emph{modular} structure inherent to bimanual tasks.

A \emph{modular} perspective on bimanual manipulation is supported by neuroscience: human bimanual manipulation is the coordination of arm-specific motor primitives rather than a single monolithic controller. Dedicated neural circuits, such as the Supplementary Motor Area (SMA) and the corpus callosum, orchestrate and synchronize the two arms~\citep{Sadato9667, Swinnen2002}. Similar principles have proven effective in vision-language modeling, where interaction between modality-specific backbones improves its efficiency and effectiveness~\citep{liang2024mixtureoftransformerssparsescalablearchitecture}.

\begin{figure}[t]
    \centering
    \includegraphics[width=0.95\textwidth]{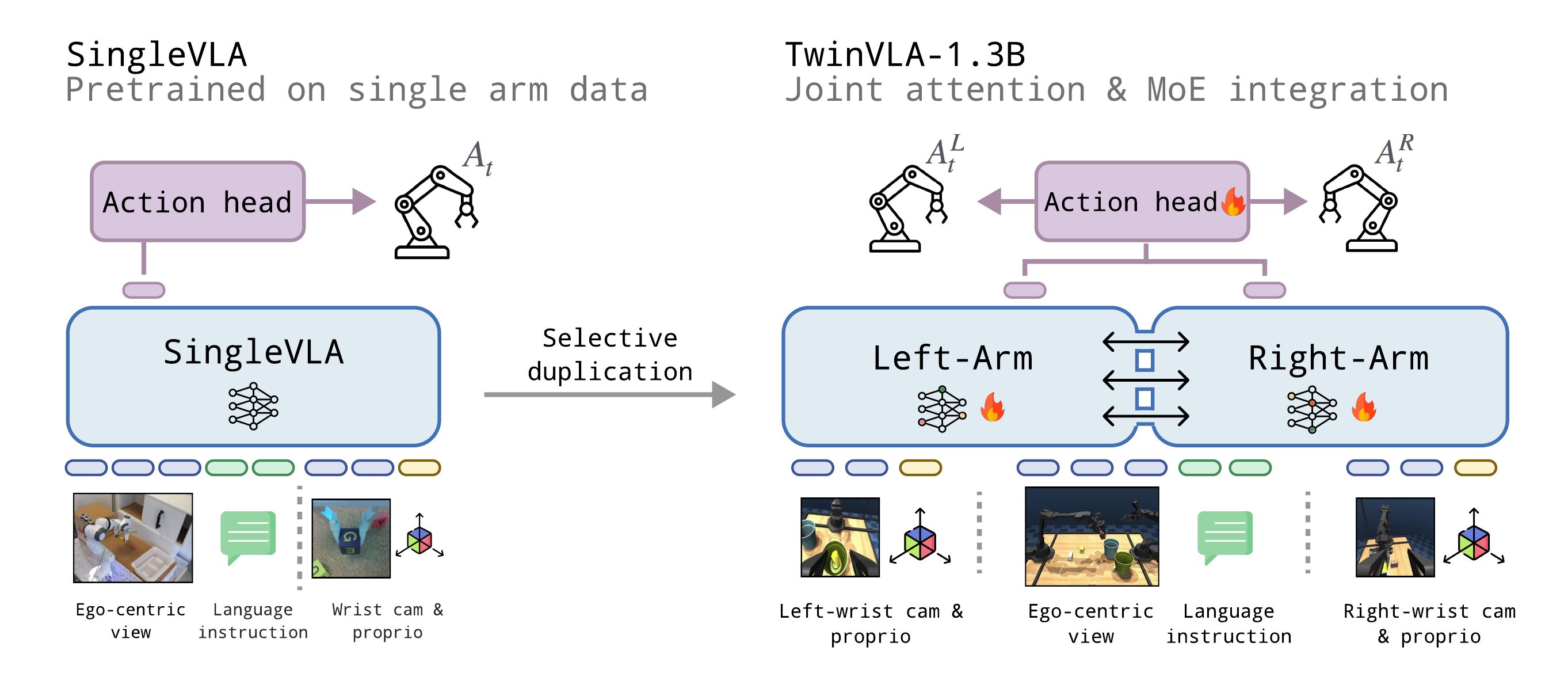}
    \vspace{-1em}
    \caption{\textbf{Overview of TwinVLA.} Inspired by humans' two-arm coordination for bimanual manipulation, TwinVLA duplicates a VLM backbone pretrained on cross-embodiment single-arm data (\textit{Left}) to form two arm-specific branches linked via \textbf{Joint Attention }(\textit{Right}). Shared inputs (ego-centric views, language instructions) are routed via a mixture-of-experts (MoE) to improve computational efficiency. Only the VLM backbone is duplicated, keeping the increase in model size minimal.}
    \label{fig:twinvla}
\end{figure}

Inspired by these insights, we propose TwinVLA, a modular architecture that operationalizes this coordination-centric view. Instead of training from scratch, TwinVLA leverages a pretrained single-arm VLA. Specifically, we first design a lightweight, compact single-arm VLA, which we call SingleVLA (\Cref{sec:singlevla}). We pre-train a 0.8B-size SingleVLA for single-arm manipulation on the OXE dataset~\citep{oxe2024}. We then duplicate this SingleVLA and integrate the two ``twin'' instances through a lightweight coordination method. This design is highly data-efficient: it eliminates the need for a bimanual pretraining dataset and achieves strong performance with only a small amount of bimanual demonstrations for fine-tuning.

To integrate two SingleVLAs into a bimanual policy, TwinVLA utilizes a joint attention~\citep{liang2024mixtureoftransformerssparsescalablearchitecture} across the twin models, as illustrated in \Cref{fig:twinvla}. This allows the twin SingleVLAs to exchange information and coordinate their actions, while preserving their pretrained capabilities. This approach is made feasible without significant overhead, as we duplicate only the VLM backbone and utilize a Mixture-of-Experts (MoE) to efficiently manage shared inputs. In contrast to monolithic cross-embodiment models~\citep{liu2024rdt, octo_2023, doshi2024scalingcrossembodiedlearningpolicy}, our approach yields better performance and data efficiency, significantly reducing the need for large-scale bimanual data collection and compute.

We evaluate TwinVLA across a broad range of environments, including a complex, long-horizon real-world task and a diverse suite of bimanual manipulation tasks in simulations. Despite leveraging only public single-arm data and limited bimanual fine-tuning data, TwinVLA achieves performance comparable to state-of-the-art bimanual policies. 

In summary, our main contributions are threefold: 
\begin{itemize}[leftmargin=2em,topsep=0pt,itemsep=0pt,parsep=3pt]
    \item We propose a novel modular architecture for bimanual manipulation that integrates two copies of a pretrained SingleVLA with a lightweight coordination method based on joint attention with MoE, enabling synchronized two-arm control.
    \item We present a data-efficient paradigm that adapts our twin architecture into a capable bimanual policy for a target task by fine-tuning on only a small bimanual dataset, crucially without requiring additional pretraining, thereby eliminating the need for large-scale bimanual data.
    \item Through extensive experiments across real and simulated bimanual tasks, TwinVLA matches or surpasses state-of-the-art models trained on far larger bimanual data and compute.
\end{itemize}
Together, these findings identify our modular SingleVLA composition approach as a scalable, efficient path to high-performance bimanual manipulation.

\begin{figure}[t]
    \centering
    \begin{subfigure}[t]{0.43\textwidth}
        \centering
        \includegraphics[width=\textwidth]{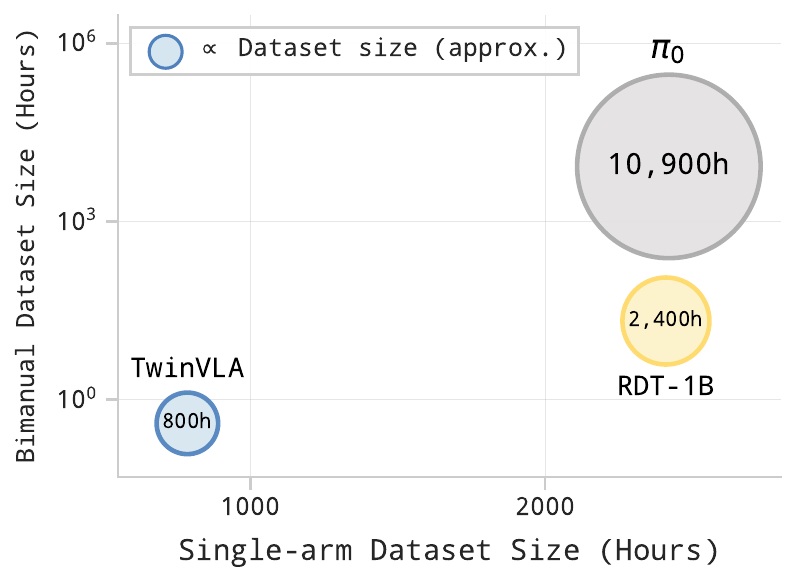}
        \caption{Data efficiency comparison}
    \end{subfigure}
    \begin{subfigure}[t]{0.45\textwidth}
        \centering
        \includegraphics[width=\textwidth]{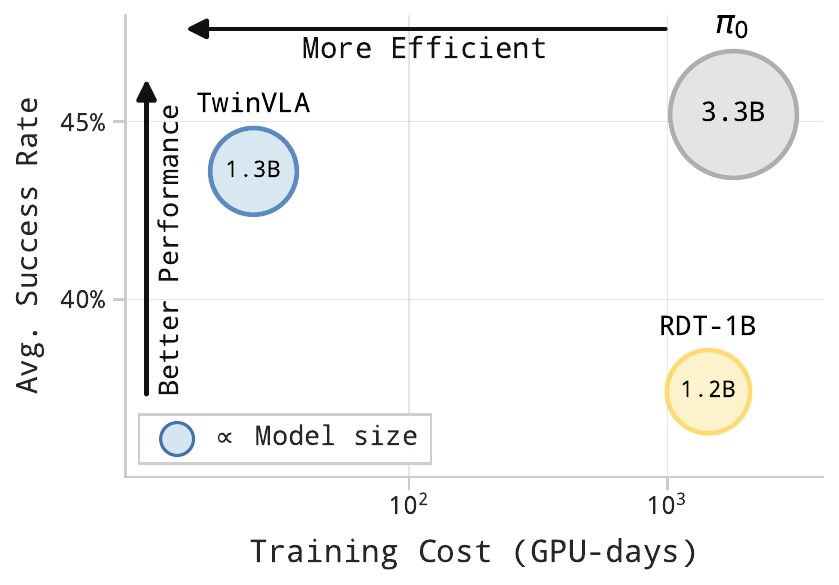}
        \caption{Compute efficiency comparison}
    \end{subfigure}
    \hfill
    \caption{\textbf{(a) Data efficiency.}~TwinVLA requires only $\sim800$h of single-arm and $50$ episodes of target bimanual data, significantly less than RDT-1B ($\sim2,400$h) and $\pi_0$ ($\sim10,900$h) in total. \textbf{(b) Compute efficiency.}~RDT-1B and $\pi_0$ require high compute (exceeding $1,000$ H100 GPU-days), whereas TwinVLA achieves higher or comparable performance with only $25$ H100 GPU-days.}
    \label{fig:twinvla_comparison}
    \vspace{-1em}
\end{figure}

\section{Related Work}

Bimanual manipulation policies are essential to enable robots to perform complex tasks that require coordinated two-handed control, such as folding laundry~\citep{bersch2011bimanual, avigal2022speedfolding}, assembling parts~\citep{stavridis2018bimanual}, or wiping the plate~\citep{black2025pi, chi2024universalmanipulationinterfaceinthewild}. Learning effective bimanual policies is challenging due to high-dimensional, tightly coupled action spaces and the scarcity of high-quality bimanual demonstrations~\citep{lee2020learning, xie2020deepimitationlearningbimanual}. Consequently, specialist methods, such as Diffusion policy~\citep{chi2023diffusion} and ACT~\citep{zhao2023learning}, trained only on target-task demonstrations, struggle on precise, long-horizon tasks.

Recent works have explored various architectures for bimanual control to explicitly model the inter-dependencies between arms~\citep{lee2024interact, kobayashi2025bivlabilateralcontrolbasedimitation}, or focus on high-level language planning via VLM~\citep{10831380}. Anybimanual~\citep{lu2025anybimanualtransferringunimanualpolicy} introduces a high-level skill manager to coordinate primitives and visual aligner to mask 3D voxels for decoupled policies, benefiting from both high-level managing and architectural inductive bias. While promising, it is difficult to generalize these methods, as they are often limited to small-scale scenarios, handle only low-dexterity tasks, or backbone constraints~\citep{DBLP:journals/corr/abs-2407-00278, shridhar2022peract}.

Alternatively, another line of research extends successful unimanual Vision-Language-Action (VLA) models~\citep{liu2023visualinstructiontuning, rt22023arxiv, li2024visionlanguagefoundationmodelseffective} to bimanual tasks. This transition is challenging due to the scarcity of bimanual data, as public datasets are predominantly unimanual. To overcome this, prior work trains `monolithic' models, requiring large-scale bimanual data collection and intensive pretraining. For example, RDT-1B~\citep{liu2024rdt} required massive pretraining and fine-tuning (reportedly a month on $48$ H100 GPUs), and $\pi_0$~\citep{black2024pi_0} relies on a $10,000$-hour proprietary dataset, both incurring high computational costs. Furthermore, the proprietary nature of these datasets limits reproducibility and broader adoption.

In contrast to both monolithic, compute-heavy pretraining and specialized architectural designs, our approach adopts a modular, coordination-centric design. While Anybimanual~\citep{lu2025anybimanualtransferringunimanualpolicy} introduces novel inductive biases for coordination, these are often difficult to integrate into general-purpose VLA frameworks due to specific backbone constraints. Our method, however, is designed to leverage and scale the existing generalist VLAs. We first train a SingleVLA on large-scale public single-arm data, duplicate to couple them, and then fine-tune it on bimanual tasks---allowing each stage to benefit from the most suitable data (see \Cref{fig:twinvla_comparison}). This composition-based approach avoids bimanual pretraining, requires only a small amount of bimanual fine-tuning, better preserves the strong capabilities of single-arm policies, and significantly improves data and compute efficiency.

\section{Preliminaries}

This paper aims to develop a data-efficient framework for learning bimanual manipulation policies by building upon pretrained single-arm Vision-Language-Action (SingleVLA) models. This section formalizes the single-arm and bimanual settings, briefly describes the VLA training objective, and introduces the core architectural concepts we leverage.

\subsection{Formulating the Bimanual VLA Policy}
\label{sec:formulation}

Our goal is to extend a pretrained SingleVLA $\pi_{\text{single}}$ into a bimanual policy $\pi_{\text{twin}}$ applicable to target bimanual tasks. A VLA $\pi(A_t \mid o_t)$ predicts an \textit{action chunk} $A_t = (a_t, a_{t+1}, \dots, a_{t+T-1})$ of length $T$ from an observation $o_t$. For single-arm manipulation, the observation $o^\text{{single}}_t=\left( (l, I_\text{ego})_t, (I_\text{wrist}, d)_t \right)$ includes a language prompt $l$, an ego-centric image $I_\text{ego}$  (shared input), and an arm-specific wrist image $I_\text{wrist}$ with proprioception $d$ (arm-specific input). We train $\pi_\text{single}(A_t \mid o^\text{single}_t)$ to predict the action chunk for one arm. For bimanual manipulation, the observation aggregates both right ($R$) and left ($L$) arm-specific input, $o^\text{twin}_t=\left( (l, I_\text{ego})_t, (I_\text{wrist}^{R}, d^R)_t, (I_\text{wrist}^{L}, d^L)_t \right)$, and the policy $\pi_\text{twin}(A^R_t, A^L_t \mid o^\text{twin}_t)$ outputs a joint action chunk for right and left arms.

\subsection{Training VLAs with Conditional Flow Matching}
\label{sec:training_objective}

We train our VLA models to predict continuous robot actions from observations. Each observation $o_t$ is tokenized and fed into the VLM backbone to produce an output embedding $h_t$ (from a learnable readout token $r_t$). To enable continuous action prediction from $h_t$, we attach an action head $v_\theta(A_t^\tau, h_t, d_t)$ and train it using a conditional flow matching objective. The action head is trained with the following loss function:
\begin{equation}
\mathcal{L}^T(\theta) = \mathbb{E}_{p(A_t \mid o_t), q(A_t^\tau \mid A_t)} \lVert v_\theta(A_t^\tau, h_t, d_t) - \mathbf{u}(A_t^\tau \mid A_t) \rVert^2,
\end{equation}
where $h_t$ is the VLM output embedding and $d_t$ is proprioception. This objective trains the action head $v_\theta$ to predict the reference flow $\mathbf{u}$ from a noised action chunk $A_t^\tau$ to the target action chunk $A_t$, conditioned on the VLM output and proprioception.

During inference, we sample actions using the forward Euler integration method. Starting from $A_0 \sim N(0, I)$, we iteratively update the action using the learned flow $v_\theta$:
\begin{equation}
A_t^{\tau+\delta}=A_t^\tau + \delta v_\theta(A_t^\tau, h_t, d_t),
\end{equation}
where we set the sampling step $n=10$ and use $\delta = \frac{1}{n}$.

\subsection{Mixture-Based Architectures} \label{sec:arch_preliminaries}

To adapt Transformers for multi-modal inputs, various mixture-based architectures have been explored and shown to be effective. These approaches range from combining entire, modality-specific backbones to ensembling or mixing individual layers within a single backbone. We briefly introduce two such paradigms that inform our design: a \textit{model-level} Mixture-of-Transformers (MoT), which coordinates separate backbones, and a \textit{layer-level} Mixture-of-Experts (MoE), which enables efficient, sparse computation.

The MoT architecture~\citep{liang2024mixtureoftransformerssparsescalablearchitecture} enables efficient information sharing between separate, modality-specific backbones (e.g., text and image). It introduces \textit{joint attention}, a shared self-attention layer performed over the union of multimodal inputs, allowing each modality to directly attend to the others. Meanwhile, modality-specific components such as feed-forward networks remain separate, making fusion lightweight yet effective.

MoE~\citep{shazeer2017outrageously} scales model capacity efficiently by routing each input $x$ through a weighted combination of expert feed-forward networks using a gating function, yielding $\text{MoE}(x)=\sum_i w_i E_i(x)$, where $w_i$ denotes the routing weight.

\section{TwinVLA}
\label{sec:twinvla}

TwinVLA is a modular architecture that transforms a pretrained single-arm VLA into a coordinated bimanual policy. The overall computation flow of our architecture is described in \Cref{alg:joint_attn} and \Cref{fig:moe_and_joint}. TwinVLA integrates single-arm policies through three core principles: (1) selective module duplication (\Cref{sec:duplication}), (2) cross-arm fusion via joint attention (\Cref{sec:fusion}), and (3) efficient shared representation via Mixture-of-Experts (\Cref{sec:moe}).

\begin{algorithm}[h!]
\caption{TwinVLA}
\label{alg:joint_attn}

\renewcommand{\algorithmiccomment}[1]{\hfill\textcolor{gray}{\footnotesize $\triangleright$ #1}}

\begin{algorithmic}

\State $X^m_0$: encoded inputs from $o_t^\text{twin}$ (\Cref{sec:formulation}) for each input $m \in \{\text{shared}, \text{left}, \text{right}\}$.
\State $\text{FFN}_b$: feed-forward network layer from each backbone $b \in \{\text{left}, \text{right}\}$.
\State $N$: Number of transformer layers
\For{$n=0$ to $N-1$} \Comment{Iterate every transformer layer}
\State {\color{gray} // Prepare Q, K, V for each input $m$}
\For{each input $m \in \{\text{shared}, \text{left}, \text{right}\}$}
    \State $Q^m_n, K^m_n, V^m_n \leftarrow {\color{projColor}\text{Norm}(\text{Proj}(X^m_n))}$  \Comment{Input-specific projections, \Cref{alg:arith_alg}}
\EndFor
\State {\color{gray} // Joint attention across inputs with attention re-weighting}
\State {$\{A^m_n\} \leftarrow {\color{jointAttnColor} \text{JointAttention}(\{Q^m_n\}, \{K^m_n\}, \{V^m_n\}, M)}$} \Comment{\Cref{alg:joint_attn_alg}, with mask $\color{jointAttnColor}M$~\Cref{fig:joint_right}}
\State {\color{gray} // Residual \& FFN / MoE}
\For{each input $m \in \{\text{shared}, \text{left}, \text{right}\}$} 
    \State $H^m_n \leftarrow X^m_n + {\color{projColor}\text{Norm}(\text{Proj}(A^m_n))}$ \Comment{Input-specific output projection, \Cref{alg:arith_alg}}
    \State {$F^m_n \leftarrow {\color{moeColor} \text{MoE}(H^m_n)}$} \textbf{if } $m=\text{shared}$ \textbf{ else} $\text{FFN}_m(H^m_n)$ \Comment{MoE for shared input, \Cref{eq:moe_definition}}
    \State $X^m_{n+1} \leftarrow H^m_n + {\color{projColor}\text{Norm}(F^m_n)}$ \Comment{Residual connection with norm, \Cref{alg:arith_alg}}
\EndFor

\EndFor
\State \textbf{return} $\{X^m_N\}$ \Comment{Return outputs, this will be used for action decoding}
\end{algorithmic}
\end{algorithm}

\subsection{Single-Arm Policy Duplication} 
\label{sec:duplication}

We first pre-train a VLA on a single-arm dataset, which we refer to as SingleVLA. Note that existing pre-trained models can also be used for this purpose. To construct TwinVLA from SingleVLA, we initialize the twin policies for the left and right arms by copying the pretrained SingleVLA. However, instead of duplicating the full model, we share the vision encoder and DiT~\citep{Peebles2022ScalableDM} action head while fully replicating the VLM. Each arm has its own lightweight proprioception encoder. This design yields a compact $1.3$B-parameter model, comparable to the $1.2$B-parameter RDT-1B, without significantly increasing computational cost.

Visual inputs are processed by the shared encoder, and each VLM produces readout tokens that are jointly decoded by the shared DiT. This design is motivated by the principle that general visual understanding (image encoding) and low-level motor control (action decoding) are largely embodiment-agnostic skills that can be effectively shared for both arms. In contrast, the VLM, which decides output action given encoded observation, is fully replicated to allow for specialized control.

\subsection{Joint Attention for Cross-arm Fusion}
\label{sec:fusion}

\begin{figure}[h]
    \centering
    \begin{subfigure}[t]{0.4\textwidth}
        \centering
        \includegraphics[width=0.9\textwidth]{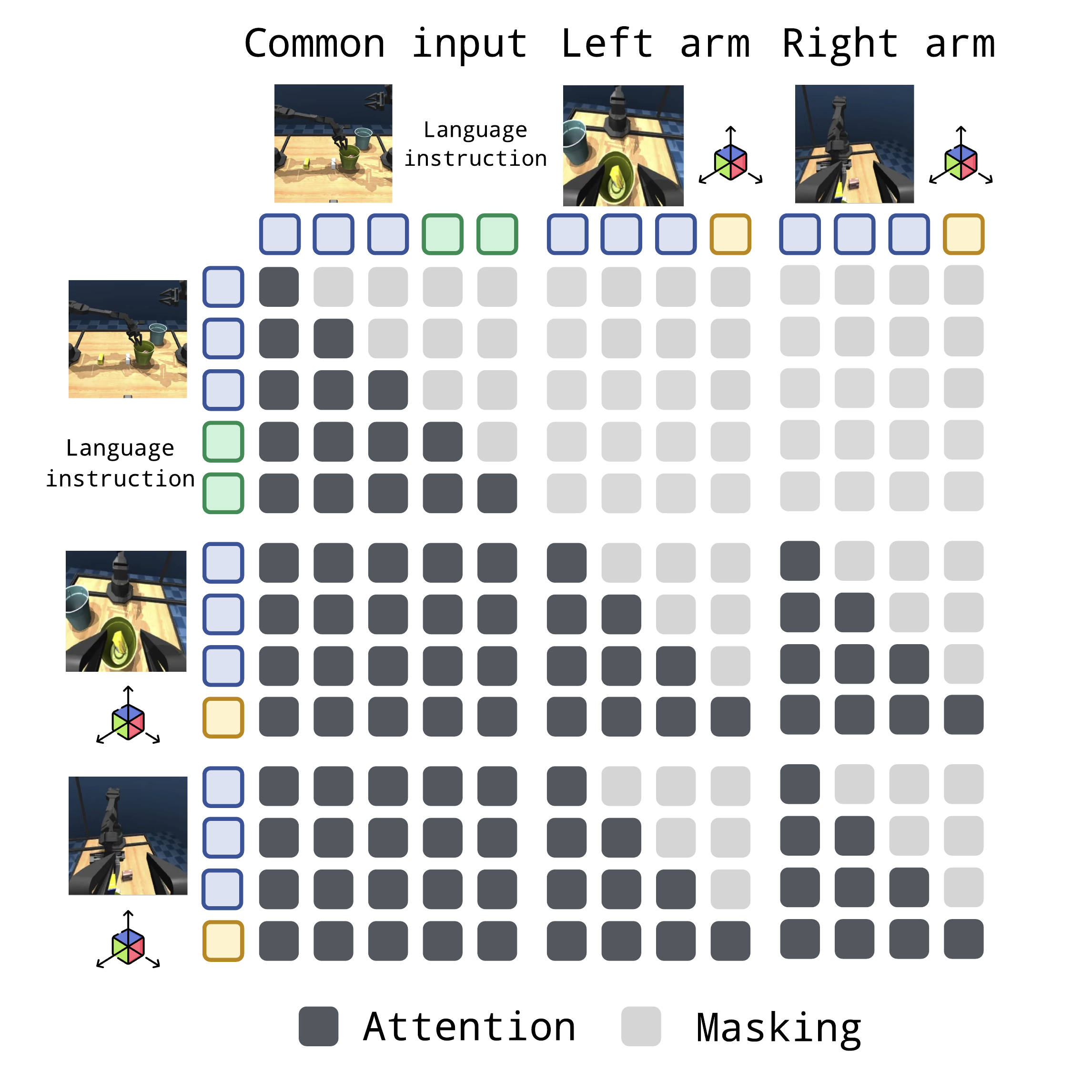}
        \caption{Causal joint attention mask}
        \label{fig:joint_right}
    \end{subfigure}
    \begin{subfigure}[t]{0.54\textwidth}
        \centering
        \includegraphics[width=0.9\textwidth]{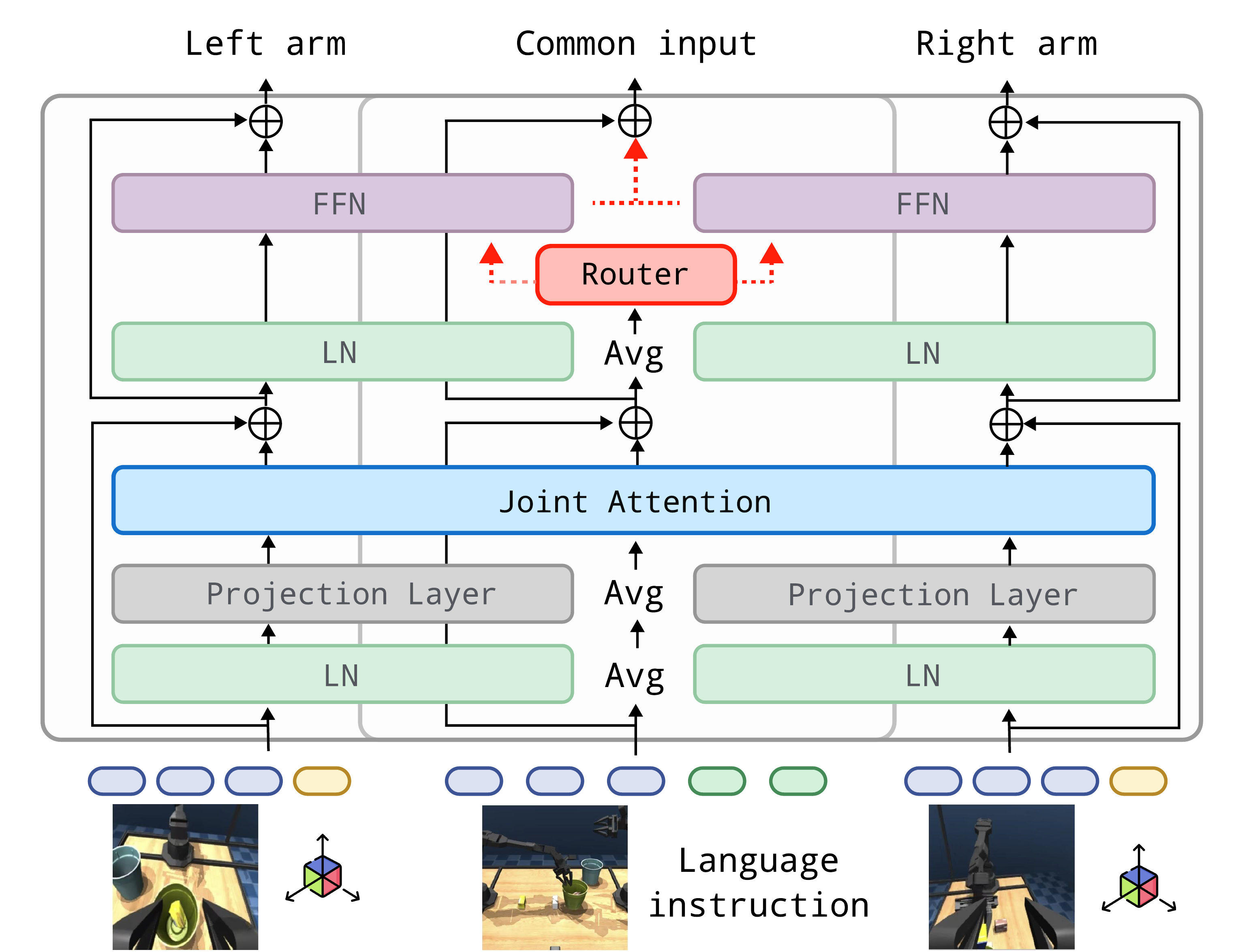}
        \caption{Transformer block of TwinVLA}
        \label{fig:moe_left}
    \end{subfigure}
    \hfill
    \caption{\textbf{(a) Causal attention mask for joint attention.} It preserves causality while processing shared, left, and right inputs in parallel. \textbf{(b) TwinVLA joint attention mechanism.} The two VLMs share information, and the shared modality $(l, I_\text{ego})_t$ is further processed by MoE to more efficiently leverage both VLMs.}
    \label{fig:moe_and_joint}
\end{figure}

We integrate arm-specific inputs using a {\color{jointAttnColor}Joint Attention} mechanism inspired by MoT~\citep{liang2024mixtureoftransformerssparsescalablearchitecture}. As illustrated in \Cref{fig:moe_left} and \Cref{alg:joint_attn}, this is achieved by sharing only the self-attention layers across the VLM backbones. Specifically, we concatenate the $\color{jointAttnColor}Q, K, V$ from both backbones, perform self-attention, and subsequently split the outputs back to their respective streams, while other components such as projections use arm-specific networks from each arm's VLM backbone. Unlike $\pi_0$~\citep{black2024pi_0}, which links a VLM with an action head, we connect two VLMs directly. We elaborate joint attention mechanism in detail on~\Cref{alg:joint_attn_alg}.

\textbf{Causal joint attention mask.} Effective joint attention requires appropriate attention masking. Standard LLMs use a lower-triangular attention mask for causal prediction. To support joint attention among the shared and arm-specific inputs, we designed the attention mask for TwinVLA as shown in~\Cref{fig:joint_right}. Specifically, we embed lower-triangular masks within each arm’s region while treating the shared modality as fully accessible. Each arm also attends to half of the other’s tokens, enabling symmetric cross-arm interaction without violating autoregressive constraints.

\subsection{Mixture-of-Experts Integration}
\label{sec:moe}

In TwinVLA, feeding shared inputs $(l, I_\text{ego})_t$ redundantly to both VLMs significantly increases VRAM usage. To address this, we process shared tokens as a single sequence by employing a {\color{moeColor}MoE} mechanism that dynamically routes shared tokens between the two VLM experts:
\begin{equation}
\label{eq:moe_definition}
{\color{moeColor}\text{MoE}(x)} = w_{\text{left}} \cdot \text{FFN}_{\text{left}}(x) + (1 - w_{\text{left}}) \cdot \text{FFN}_{\text{right}}(x).
\end{equation}
For calculating $w_\text{left}$, we add a linear layer that takes the embedding as input and outputs the weights via a softmax function. For other components like {\color{projColor}$\text{Projection},\text{LayerNorm}$}, we implement an output-averaging strategy inspired by task arithmetic~\citep{10.5555/3692070.3694018}. By processing inputs through both backbones and averaging their outputs, we functionally simulate a shared layer without physically merging parameters (see \Cref{fig:moe_left} center). This efficient design reduces VRAM usage by 21\%, enabling training with a batch size of $8$ on a single $40$ GB GPU. 

\textbf{Attention re-weighting.} A potential side effect of introducing new arm-specific tokens is that the model's learned attention patterns can be disrupted, shifting focus away from the pretrained shared modalities. To mitigate this and preserve the valuable pretrained knowledge, we re-scale the attention scores for the shared modality (\Cref{alg:attention_reweighting_simple}). This maintains pretrained modality importance, allowing the model to bypass an initial adaptation phase and focus directly on the target task—a benefit evidenced by a lower initial loss and converged loss during fine-tuning.

\section{Experiments}

In this paper, we propose TwinVLA to achieve strong bimanual manipulation performance with minimal bimanual data by fully leveraging a single-arm VLA pretrained on abundant single-arm data. Our empirical studies aim to answer the following questions:

\begin{itemize}[leftmargin=2em,topsep=0pt,itemsep=0pt,parsep=3pt]
    \item How does TwinVLA compare to state-of-the-art methods across diverse bimanual tasks, without any bimanual pretraining (\Cref{sec:real_exp,sec:sim_exp})? 
    \item How quickly can TwinVLA adapt to new bimanual tasks (\Cref{sec:data-efficiency})? 
    \item Does TwinVLA retain core VLA properties---language-following and robustness to unseen scenes and instructions (\Cref{sec:multi,sec:robustness})? 
    \item How much does each key design choice contribute to overall performance (\Cref{sec:ablation})?
\end{itemize}

\begin{figure}[t]
    \centering
    \begin{subfigure}[t]{0.49\textwidth}
        \centering
        \includegraphics[height=3.9cm]{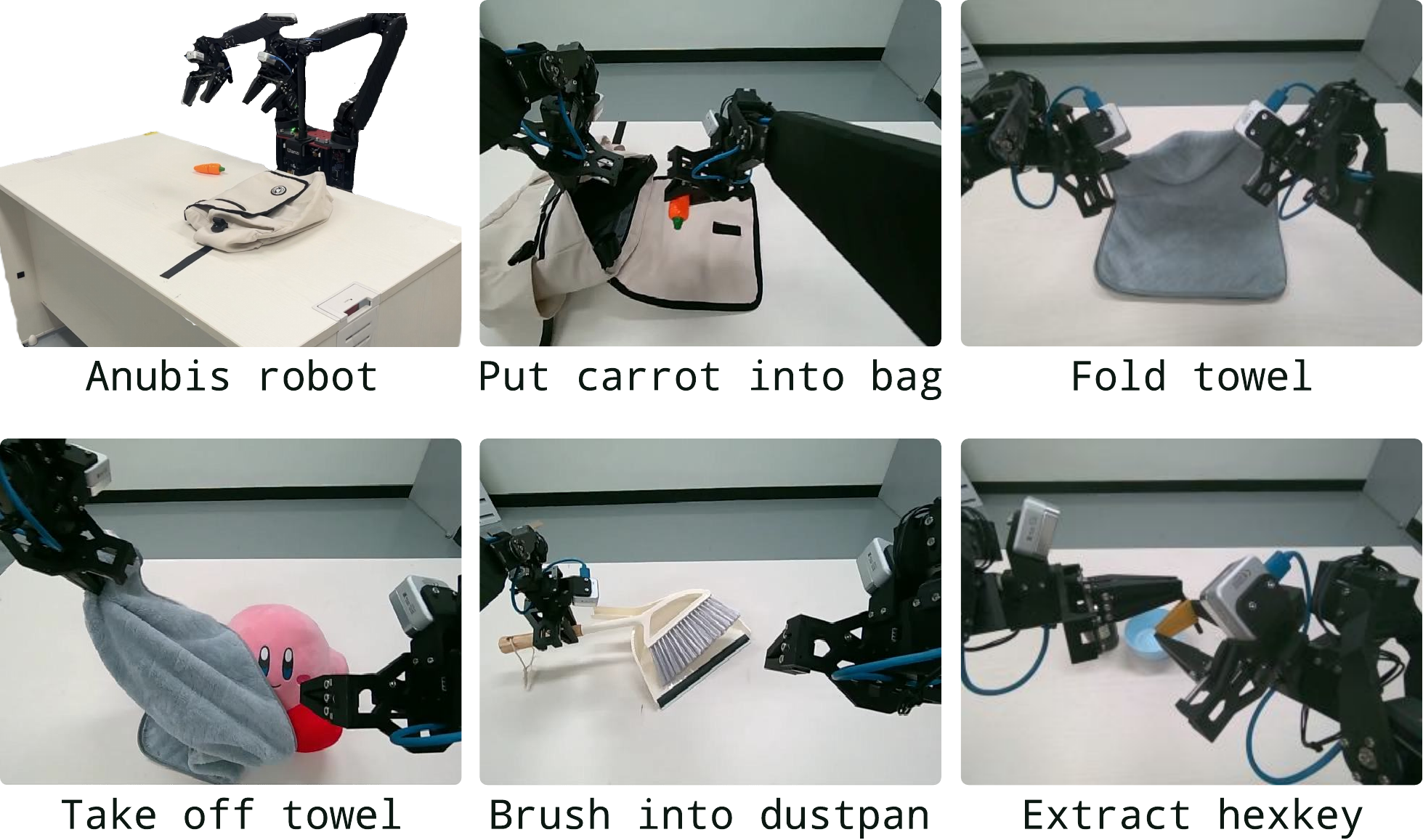}
        \caption{Real-world tasks}
        \label{fig:eval:real}
    \end{subfigure}
    \hfill
    \begin{subfigure}[t]{0.49\textwidth}
        \centering
        \includegraphics[height=3.9cm]{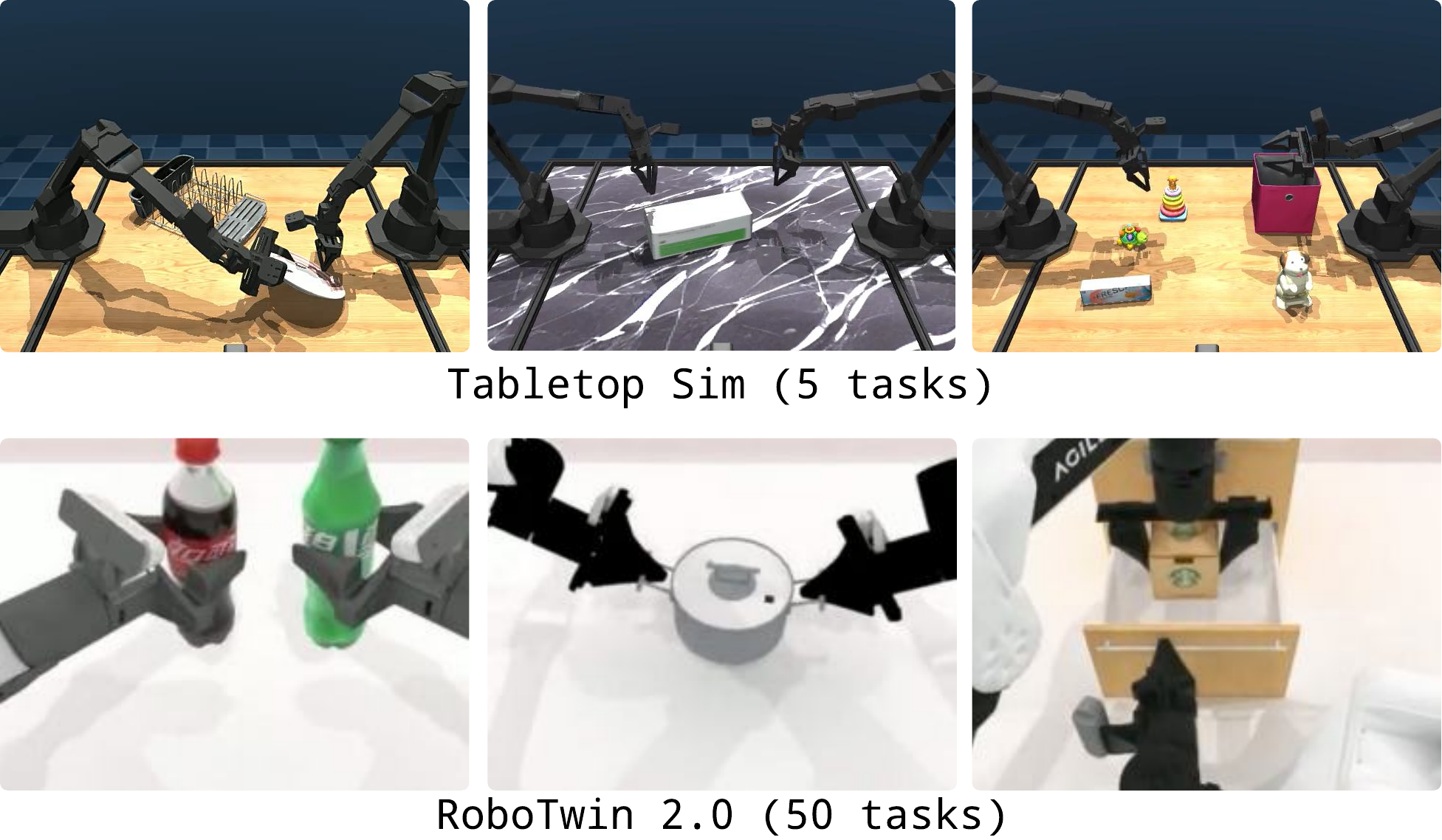}
        \caption{Simulation tasks}
        \label{fig:eval:sim}
    \end{subfigure}
    \vspace{-0.5em}
    \caption{\textbf{Experimental setups.} (a)~We evaluate TwinVLA on five real-world bimanual tasks using an Anubis robot. (b) We further analyze TwinVLA on a large suite of simulation tasks: $5$ tasks in Tabletop-Sim and $50$ tasks in RoboTwin 2.0.}
    \label{fig:eval_overview}
\end{figure}

\subsection{Compared Methods}

We evaluate TwinVLA against three bimanual manipulation policies, each representing a different point in the design space. 

\begin{itemize}[leftmargin=2em,topsep=0pt,itemsep=0pt,parsep=3pt]
    \item \textbf{RDT-1B}~\citep{liu2024rdt}: This serves as our direct baseline. With a comparable size($1.2$B vs. TwinVLA's $1.3$B parameters), it represents the standard monolithic approach that requires substantially larger resources ($\sim$2,400h data, $\sim$$1$,$440$ H100 days vs. $\sim$800h single-arm data,  $\sim$$25$ H100 days).
    \item $\boldsymbol{\pi_0}$~\citep{black2024pi_0}: We include this as an upper-bound, as this is $3.3$B-parameter VLA trained on over $10$K hours of proprietary robot data. Our goal is to assess how closely TwinVLA can approach this performance ceiling with far greater efficiency.
    \item \textbf{Diffusion Policy (DP)}~\citep{chi2023diffusion}: This is a strong baseline method in low-data regime with $271$M parameters, used to demonstrate the crucial benefits of pretraining.
\end{itemize}

\begin{figure}[h]
  \centering
  \vspace{-0.5em}
  \includegraphics[width=\textwidth]{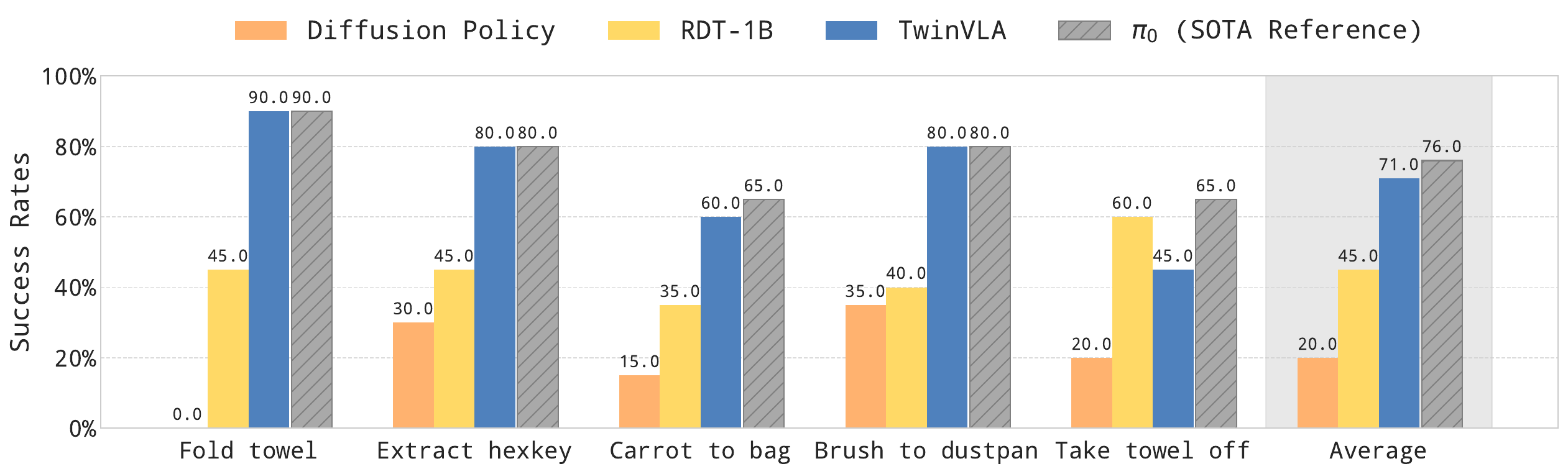}
  \vspace{-0.5em}
  \caption{\textbf{Success rates on real-world tasks.} TwinVLA outperforms RDT-1B and DP on average. Moreover, TwinVLA shows comparable performance with $\pi_0$ while trained only on target data.}
  \label{fig:real-results}
  \vspace{-1.0em}
\end{figure}

\subsection{Real-World Experiments}
\label{sec:real_exp}

\textbf{Environment.} For real-world experiments, we use a dual-arm robot, Anubis~\citep{anubis2025}, as shown in \Cref{fig:eval:real}. Anubis has two $6$ DoF arms with parallel-jaw grippers. The robot is equipped with two wrist-mounted cameras and a single ego-centric view camera. 

\textbf{Tasks.} We design five long-horizon tabletop manipulation tasks, which require careful coordination and accurate motions: \texttt{Fold towel}, \texttt{Extract hexkey}, \texttt{Carrot to bag}, \texttt{Brush to dustpan}, and \texttt{Take towel off}, and one task set, \texttt{Put X into pot}. We collect $50$ episodes for each task using absolute EEF control. Each method is fine-tuned for each task and evaluated with $20$ rollouts.

\textbf{Results.} As presented in \Cref{fig:real-results}, our model, TwinVLA, significantly outperforms RDT-1B. This achievement is remarkable considering the data disparity: TwinVLA is pretrained on just $\sim$800h of single-arm data, in contrast to RDT-1B's usage of a $\sim$2,400h dataset mixed with bimanual trajectories, which highlights the data efficiency of our approach. While DP's low performance confirms the necessity of pretraining, $\pi_0$ achieved the highest overall performance with significantly higher costs.

\subsection{Simulation Experiments}
\label{sec:sim_exp}

\textbf{RoboTwin 2.0.} We use the RoboTwin 2.0 benchmark~\citep{chen2025robotwin}, consisting of $50$ bimanual tasks. Adhering to the official evaluation protocol, we fine-tune a model per task with $50$ generated demonstrations and perform $100$ test rollouts under both ``Easy'' and ``Hard'' settings. For Easy tasks, test scenes match the training data, but the instructions are novel. The Hard tasks introduce variations in texture, object position, and height. For compared methods, we use the results reported from RoboTwin 2.0~\citep{chen2025robotwin}.

\textbf{Tabletop-Sim.} To assess dexterous scenarios beyond tasks in RoboTwin, we develop Tabletop-Sim\footnote{Our simulation setup is similar to the concurrent work Aloha-Sim, released by Google DeepMind~\citep{aloha_sim_github}.}, a tabletop simulation environment based on \texttt{dm\_control}~\citep{tunyasuvunakool2020} and assets from ALOHA2~\citep{aloha2team2024aloha2enhancedlowcost} and GSO object dataset~\citep{DBLP:journals/corr/abs-2204-11918}. We design $5$ representative tasks that require precise bimanual coordination. Specifically, we define four single-tasks and one multi-task: \texttt{dish-drainer}, \texttt{handover-box}, \texttt{shoes-table}, \texttt{lift-box}, and \texttt{put X box into Y pot}. In the ``Hard'' tasks, we vary background textures and objects. We collect $50$ episodes on each task using absolute EEF control, and fine-tune a model per task, and perform $500$ evaluation rollouts for both ``Easy'' and ``Hard'' settings.

\begin{figure}[t]
    \centering
    \includegraphics[width=\textwidth]{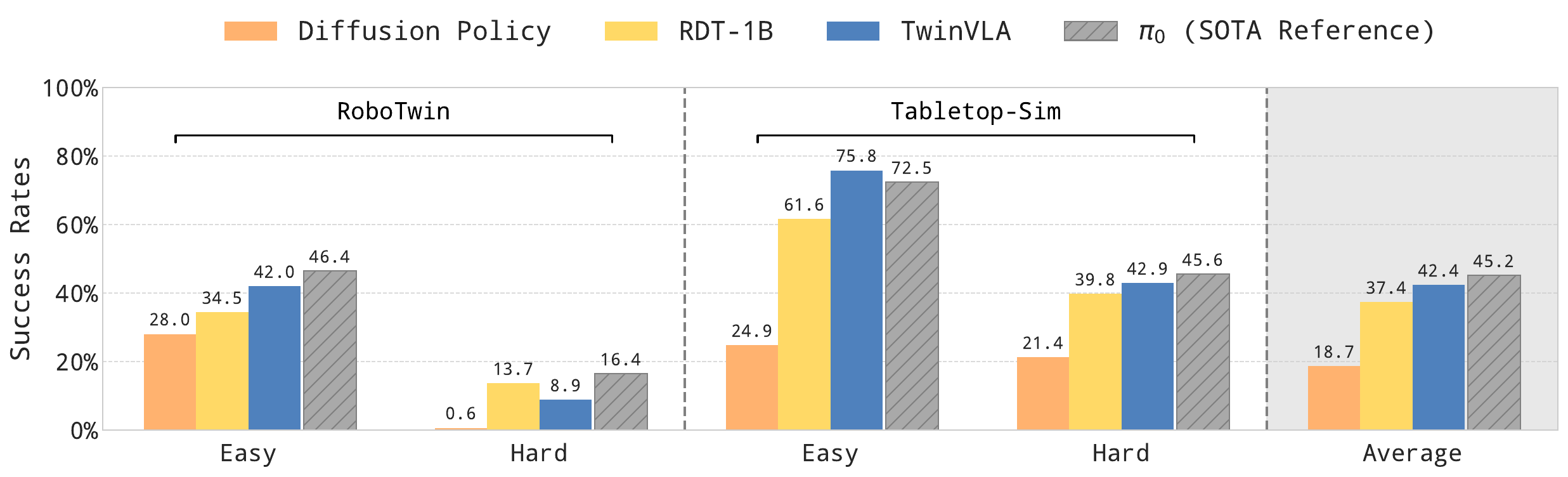}
    \vspace{-1.0em}
    \caption{\textbf{Average success rates for diverse bimanual tasks.} Despite being pretrained solely on single-arm datasets, \textbf{TwinVLA} outperforms other methods except \textbf{$\pi_0$}.}
    \label{fig:eval_results}
\end{figure}

\textbf{Results.} The results in \Cref{fig:eval_results} show the average success rates of TwinVLA and compared methods. DP, trained from scratch, shows the worst performance, highlighting the importance of pretraining. Once again, we observe that TwinVLA outperforms RDT-1B in most scenarios, except for the RoboTwin Hard tasks, and achieves comparable performance with $\pi_0$ by effectively leveraging single-arm data and modularity of bimanual manipulation. Notably, in Tabletop-Sim Easy tasks, TwinVLA even outperforms $\pi_0$, which is trained on an extensive corpus of high-quality bimanual pretraining data. This demonstrates TwinVLA's advantages in scenarios demanding higher dexterity and significant bimanual coordination.

\subsection{Data Efficiency}
\label{sec:data-efficiency}

\begin{wrapfigure}{r}{0.35\textwidth}
  \vspace{-5.0em}
  \centering
  \includegraphics[height=4.2cm]{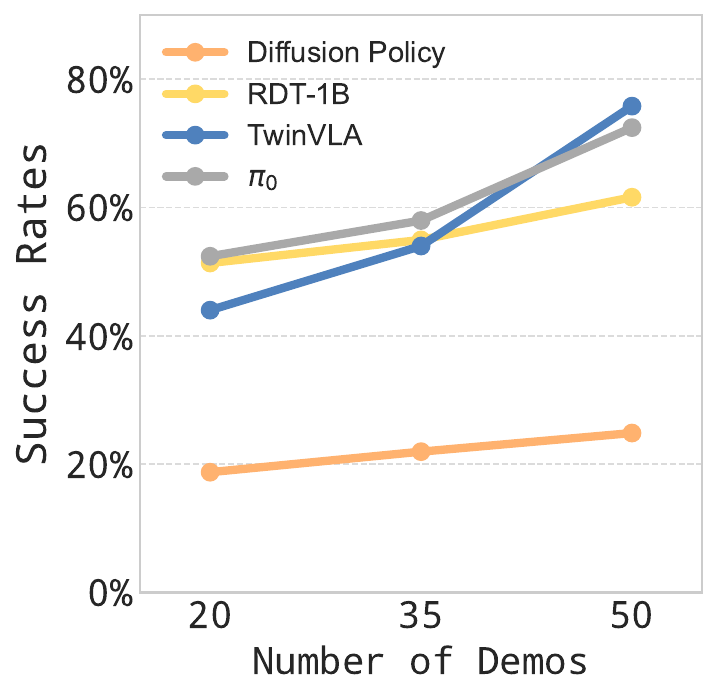}
  \vspace{-1em}
  \caption{\textbf{Average success rates on the Tabletop-Sim Easy tasks.} Models are evaluated after fine-tuning with $20$, $35$, and $50$ demonstrations.}
  \label{fig:efficiency-plot}
  \vspace{-1em}
\end{wrapfigure}

TwinVLA exhibits data efficiency in two key aspects: pretraining and fine-tuning. For pretraining, it is efficient because it does not require supplemental bimanual data. For fine-tuning, it learns new tasks rapidly because its structural inductive bias facilitates the efficient transfer and application of its pretrained single-arm knowledge. We validate this efficiency in Tabletop-Sim Easy environment, comparing model's average success rates with varying amounts of demonstration data. As illustrated in \Cref{fig:efficiency-plot}, TwinVLA exhibits a steep learning curve. Despite a modest start with $20$ demonstrations, it quickly surpasses the performance of RDT with just $50$ demonstrations, highlighting its exceptional data efficiency.

\subsection{Policy Robustness}
\label{sec:robustness}

One of the advantages of VLAs is their robustness to unseen situations and novel language instructions, thanks to pretraining. As shown in \Cref{fig:eval_results}, TwinVLA outperforms RDT-1B by $3.3\%$ even in the Hard setup of Tabletop-Sim, which involves different textures and objects. 

\begin{wraptable}{r}{0.35\textwidth}
    \centering
    \scriptsize
    \vspace{-10pt} 
    \caption{Comparison of success rates for the \texttt{Fold towel} task in challenging scenes.} 
    \label{tab:real_rob}
    \begin{tabular}{lcc} 
        \toprule
        Model & Low light & With distractors \\
        \midrule
        RDT & 15.0\% & 15.0\% \\
        $\pi_0$ & 40.0\% & \textbf{60.0\%} \\
        \textbf{TwinVLA} & \textbf{45.0\%} & 25.0\% \\
        \bottomrule
    \end{tabular}
    \vspace{-10pt}
\end{wraptable}

The RoboTwin benchmark, both in the Easy and Hard setups, uses evaluation language instructions that are unseen during training. Here, TwinVLA again shows $7.48\%$ better performance than RDT-1B in the Easy setup. Although TwinVLA's performance on the RoboTwin Hard tasks is $3.72\%$ lower than that of RDT-1B, it still outperforms a non-pretrained Diffusion policy by $9.38\%$. This result demonstrates that TwinVLA possesses sufficient robustness as a bimanual VLA, even without being pretrained on large-scale bimanual manipulation data.

In \Cref{tab:real_rob}, we additionally compared success rates in unseen real-world settings (see \Cref{fig:fold-rob})—specifically low-light and distractor-heavy environments—using the \texttt{Fold towel} task. TwinVLA is robust to lighting changes but less effective with distractors. Meanwhile, $\pi_0$ works robustly in both cases, and RDT-1B achieves the lowest success rates.

\subsection{Language Following Evaluations}
\label{sec:multi}

\begin{figure}[h]
    \vspace{-0.5em}
    \centering
    \begin{subfigure}[b]{0.30\textwidth}
        \centering
        \includegraphics[height=4.30cm]{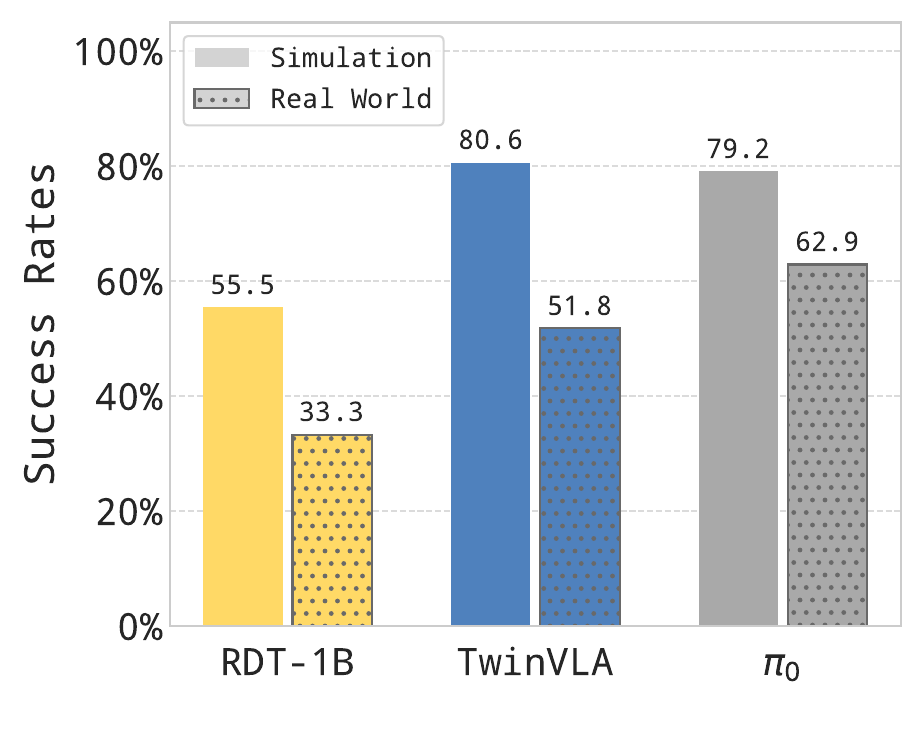}
        \caption{Language task results}
        \label{fig:multi-task-results}
    \end{subfigure} 
    \hfill
    \begin{subfigure}[b]{0.68\textwidth}
        \centering
        \includegraphics[height=4.30cm]{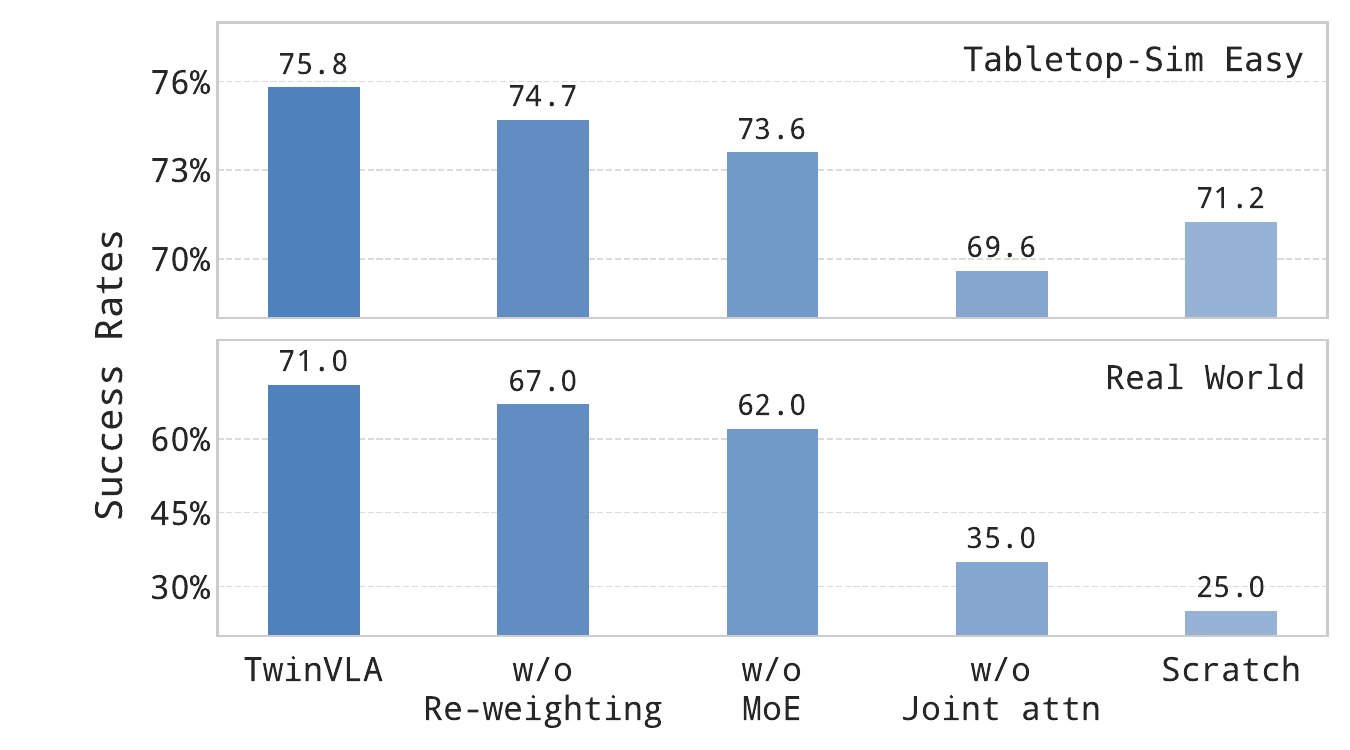}
        \caption{Ablation results}
        \label{fig:ablation-plot}
    \end{subfigure}
    \caption{\textbf{Language following task and ablation results.} (a) We evaluate average success rates on the language following tasks in the real world and Tabletop-Sim. (b) Ablation studies in the real world and Tabletop-Sim Easy tasks.}
\end{figure}

A known challenge is that fine-tuning VLMs on robotic data can degrade their ability to faithfully follow nuanced instructions. We therefore evaluate how effectively our model preserves this core capability in a multi-task setting. We evaluated the ``\texttt{Put X into pot}'' task across both simulation and real-world settings. As observed in~\Cref{fig:multi-task-results}, TwinVLA outperforms both RDT-1B and $\pi_0$. We believe this performance stems from effectively preserving the knowledge acquired during single-arm pretraining through careful fine-tuning.

\subsection{Ablations}
\label{sec:ablation}

In this section, we conduct a sequential ablation study to analyze the cumulative impact of our key design choices on performance. Starting from the full TwinVLA model, we progressively remove each component in a specific order: first Attention Re-weighting, followed by MoE integration, and finally Joint Attention. This method reveals how performance degrades as each component of our architecture is stripped away. The results on our real-world and Tabletop-Sim Easy tasks are reported in \Cref{fig:ablation-plot}.

\textbf{Attention re-weighting.} Removing the attention re-weighting mechanism (\textit{w/o Re-weighting}) increased the initial fine-tuning loss by $\textbf{40\%}$ and decreased final performance by $\textbf{1.1\%}$ and $\textbf{4.0\%}$ in simulation and real world, respectively. This demonstrates that our re-weighting strategy successfully mitigates the input distribution shift between pretraining and fine-tuning. 

\textbf{MoE integration.} Building on the previous ablation, we next remove the MoE integration (\textit{w/o MoE}). This additional change increased the token sequence length by $\textbf{28\%}$ and increased VRAM usage by $\textbf{21\%}$, making VLA training more burdensome. Surprisingly, it also further decreases the success rate by \textbf{1.1\%} and \textbf{5.0\%}, suggesting that MoE integration eliminates redundant processing of shared inputs while maintaining the performance.

\textbf{Joint attention.} Lastly, removing the joint attention mechanism (\textit{w/o Joint attn}) causes the most significant additional performance drops of $\textbf{4.0\%}$ and $\textbf{27.0\%}$ in simulation and the real world, respectively. This impact is particularly pronounced in real-world tightly coupled bimanual tasks, confirming that joint attention is a critical mechanism for bimanual coordination.

\textbf{Effect of single-arm pretraining.} As a separate, foundational experiment, we assess the role of pretraining by training a model from scratch without OXE dataset (\textit{Scratch}). This resulted in a $\textbf{4.6\%}$ performance drop in simulation and a stark $\textbf{46.0\%}$ in real world. This result confirms that effective cross-arm coordination is essential for bimanual manipulation and validates joint attention as the critical mechanism for achieving it in our model.

\textbf{Twin structure.} While we have confirmed that joint attention effectively connects the two modules, a crucial question remains: how does this approach compare to a monolithic model that is inherently unified from the start? To answer this, we revisit our comparison against RDT-1B, a monolithic model of a comparable 1.2B parameter size. The results are telling: TwinVLA outperforms RDT-1B by $\textbf{26.0\%}$ in the real world, $\textbf{5.0\%}$ in simulation, and $\textbf{21.8\%}$ in language-following tasks on average. This provides strong evidence that the inductive bias from the Twin Structure itself is highly beneficial for bimanual manipulation, validating our design choice over a monolithic approach.

\section{Limitations}
\label{sec:limitations}

Generalization remains limited due to the visual disparity of two arms, which differs from the single-arm pretraining distribution. Future research into mechanisms that prevent this could address data scarcity by integrating diverse data, while also improving the model explainability and the better generalization ability to unseen tasks.

Moreover, we adopt absolute end-effector (EEF) pose control, as its embodiment-agnostic nature facilitates single-arm transfer, unlike DOF-specific joint positions. Future exploration of relative absolute actions~\citep{chi2024universalmanipulationinterfaceinthewild} or shared representations could further enhance transfer efficiency.

\section{Conclusion} 
\label{sec:conclusion}

In this paper, we introduce TwinVLA, a data-efficient VLA model for bimanual manipulation. TwinVLA provides a new perspective on solving bimanual manipulation under scarce bimanual data by leveraging abundant single-arm datasets. From a small amount of bimanual demonstration data, TwinVLA learns to coordinate two copies of a SingleVLA pretrained on large-scale single-arm data via our proposed method. Through exhaustive experiments both in the real world and simulation, TwinVLA demonstrates its data-efficient learning of bimanual tasks compared to prior monolithic approaches. Beyond the bimanual setting, we believe this work serves as a blueprint for addressing inherent dataset imbalances across modalities. By illustrating how modular relationships can be exploited to bridge these data gaps, TwinVLA opens promising ways for other complex domains—such as mobile manipulation—thereby broadening the impact of large-scale robotic learning.

\subsubsection*{Acknowledgments}

This project was supported in part by Microsoft Research Asia and the Microsoft Accelerate Foundation Models Research (AFMR) grant program. This research was also supported by the Institute of Information \& Communications Technology Planning \& Evaluation (IITP) grants funded by the Korean Government (MSIT) (RS-2020-II201361, Artificial Intelligence Graduate School Program (Yonsei University); RS-2024-00436680, Global Research Support Program in the Digital Field Program), the Alchemist Project (RS-2024-00432143) funded by the Ministry of Trade, Industry \& Energy (MOTIE, Korea), and the Electronics and Telecommunications Research Institute (ETRI) grant funded by the Korean government (26ZR1100, Research on Intelligent Industrial Convergence). LG Electronics provided the Anubis robot, which was used for the experiments. The authors would like to thank Byeongjin Kang for the assistance with the preliminary experiments.

\bibliography{iclr2026_conference}
\bibliographystyle{iclr2026_conference}

\clearpage

\appendix

\section*{Appendix}
\addcontentsline{toc}{section}{Appendix}\label{appendix}

\section{SingleVLA: Efficient Single-Arm Policy Design and Pretraining}
\label{sec:singlevla}
\begin{figure}[ht]
    \centering
    \includegraphics[width=0.9\textwidth]{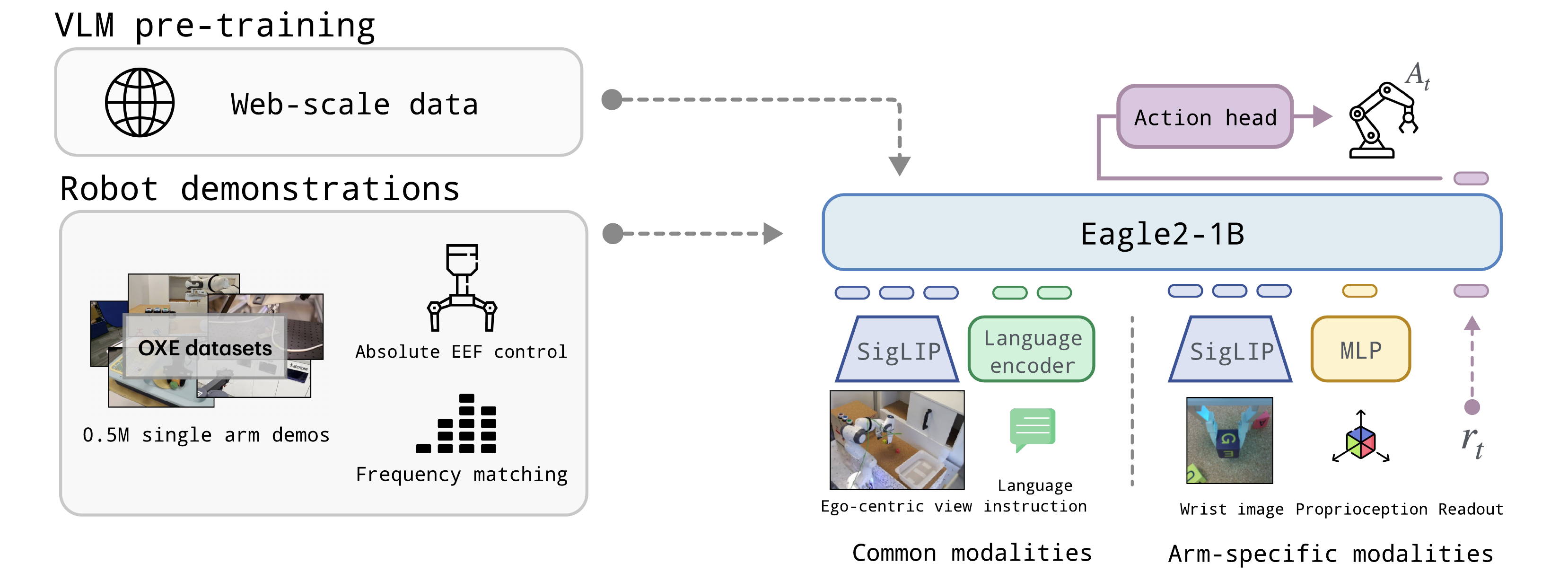}
    \caption{\textbf{Overview of SingleVLA architecture design and pretraining method.}}
    \label{fig:singlevla}
\end{figure}

This section presents the design of the SingleVLA $\pi_{\text{single}}$. While SingleVLA follows established VLA conventions, our key novelty is a duplication strategy that enables the construction of TwinVLA. Prior 7B-scale models~\citep{kim2024openvla, kim2025fine, li2024cogactfoundationalvisionlanguageactionmodel} are prohibitively large for such duplication, motivating a more efficient, lightweight Eagle2-1B~\citep{li2025eagle2buildingposttraining} based SingleVLA (Fig.~\ref{fig:singlevla}). Since we do not use language head, the overall model size became 0.8B. To acquire generalizable knowledge, we pretrain SingleVLA on a $\sim$800h subset of the OXE mix, enabling transfer across diverse environments and embodiments. Pretraining ran for \textbf{120k} steps and took about \textbf{5 days} on a cluster with \textbf{5$\times$ H100} GPUs.

To ensure effective transfer to bimanual manipulation, it is crucial to choose an appropriate \emph{action space}. Heterogeneous joint configurations across robots induce incompatible action spaces and complicate joint training. Prior work mitigates this with robot-specific decoders or high-dimensional zero-padded spaces~\citep{nvidia2025gr00tn1openfoundation, doshi2024scalingcrossembodiedlearningpolicy, octo_2023, black2024pi_0, liu2024rdt}. Instead, we convert all actions into absolute end-effector (EEF) poses, providing a consistent, semantically meaningful representation across robots that naturally extends to bimanual control. For rotation, we adopt a 6D representation~\citep{8953486}, which is well suited for neural network learning.

\subsection{Pretraining}

\begin{table}[h]
  \centering
  \caption{\textbf{SingleVLA pretraining datasets and sampling percentages.}}
  \begin{tabular}{l c}
    \toprule
    \textbf{Dataset} & \textbf{Sample Percentage} \\
    \midrule
    RT-1~\citep{rt12022arxiv} & $24.49\%$ \\
    Kuka (filtered)~\citep{Yadav_Nagar_Shah_2024} & $12.40\%$ \\
    BridgeV2~\citep{walke2023bridgedata} & $13.74\%$ \\
    Taco Play~\citep{rosete2022tacorl} & $3.10\%$ \\
    Jaco Play~\citep{dass2023jacoplay} & $0.50\%$ \\
    Viola~\citep{zhu2022viola} & $1.00\%$ \\
    Berkeley Autolab UR5~\citep{BerkeleyUR5Website} & $1.28\%$ \\
    Stanford Hydra~\citep{belkhale2023hydra} & $4.73\%$ \\
    Austin Buds~\citep{9695333} & $0.22\%$ \\
    NYU Franka Play~\citep{cui2022play} & $0.88\%$ \\
    FurnitureBench~\citep{heo2023furniturebench} & $2.40\%$ \\
    Austin Sailor~\citep{nasiriany2022sailor} & $2.33\%$ \\
    Austin Sirius~\citep{liu2022robot} & $1.84\%$ \\
    DLR EDAN (shared control)~\citep{vogel_edan_2020, quere_shared_2020} & $0.05\%$ \\
    UT Austin Mutex~\citep{shah2023mutex} & $2.38\%$ \\
    Berkeley FANUC manipulation~\citep{zhu2023fanuc} & $0.82\%$ \\
    CMU Stretch~\citep{bahl2023affordances,mendonca2023structured} & $0.16\%$ \\
    BC-Z (filtered)~\citep{jang2021bc} & $7.90\%$ \\
    FMB~\citep{doi:10.1177/02783649241276017} & $7.40\%$ \\
    Dobb-E~\citep{shafiullah2023bringing} & $1.50\%$ \\
    DROID~\citep{khazatsky2024droid} & $10.70\%$\\
    \bottomrule
  \end{tabular}
  \label{tab:dataset_table}
\end{table}

SingleVLA is pretrained on an OXE subset ($\sim$800h); dataset composition and sampling rates appear in Table~\ref{tab:dataset_table}. We adopt the dataset loader from the OpenVLA~\citep{kim2024openvla} codebase and apply sampling according to the designated weights. Because some datasets (e.g., Kuka and BC-Z) include failed trajectories, we pre-process to retain only successful ones. Regarding the action space, we convert all actions to absolute EEF control with 6D rotations. We deliberately selected an absolute representation to mitigate the error accumulation and drift issues often amplified in high-frequency bimanual control. Unlike absolute joint positions, however, absolute EEF poses preserve the embodiment-agnostic property required for heterogeneous pretraining. We define these poses relative to the robot's base frame, resulting in a 10-Dimensional action space. We further apply \emph{frequency matching} as described below.

\paragraph{Frequency matching.} Robotic datasets differ in control frequency, making fixed-length action-chunk prediction misaligned in real time. For example, a 20-step chunk spans $\sim7$ seconds in RT-1~\citep{rt12022arxiv} (3~Hz) but only $\sim1.3$ seconds in DROID~\citep{khazatsky2024droid} (15~Hz). Mixing low-frequency data like OXE~\citep{oxe2024} with high-frequency datasets can degrade pretraining quality. Inspired by $\pi_0$-FAST~\citep{pertsch2025fastefficientactiontokenization}, which uses DCT~\citep{1672377} to map $1$-second actions into a consistent space, we perform frequency matching via interpolation: all datasets are resampled to 20~Hz, improving temporal alignment and transfer to high-frequency bimanual tasks.

\subsection{Hyperparameters and Compute}

\begin{table}[h]
    \centering
    \caption{\textbf{Key hyperparameters for \textsc{TwinVLA} training.}}
    \begin{tabular}{l c c}
        \toprule
        \textbf{Hyperparameter} & \textbf{SingleVLA} & \textbf{TwinVLA}\\
        \midrule
        Global batch size & $256$ & $8$ \\
        Precision & BF16 & BF16 \\
        Gradient clipping ($L_2$) & $1.0$ & $1.0$ \\
        Learning rate & $1\times10^{-4}$ & $1\times10^{-4}$ \\
        LR scheduler & cosine & cosine \\
        Warm-up ratio & $0.01$ & $0.05$ \\
        Total steps & $120\text{k}$ & $100\text{k}$ \\
        Optimizer & AdamW & AdamW \\
        Weight decay & $1\times10^{-5}$ & $1\times10^{-5}$ \\
        Adam $\epsilon$ & $1\times10^{-8}$ & $1\times10^{-8}$ \\
        Vision backbone frozen & true & true \\
        Image augmentation & true & false \\
        Action chunk size & 20 & 20 \\
        Sampling step & 10 & 10 \\
        \bottomrule
    \end{tabular}
    \label{tab:hyperparameters}
\end{table}

Table~\ref{tab:hyperparameters} summarizes training hyperparameters for SingleVLA and TwinVLA. SingleVLA pretraining used $5$$\times$ H100 GPUs for about $5$ days. TwinVLA fine-tuning used $1$$\times$ L40S GPU for about 2 days.

\subsection{SingleVLA VLM Ablation}

We validate SingleVLA's VLM choice in the LIBERO~\citep{liu2023libero} environment using several VLMs. The LIBERO actions are converted to absolute EEF 6D control. Due to computational limits, we directly fine-tune the pretrained VLM checkpoints on LIBERO (i.e., without additional pretraining on LIBERO). Each model is evaluated with $500$ rollouts per task suite under identical random seeds. Results are shown in Table~\ref{tab:singlevla-results}.

\begin{table}[ht]
\centering
\caption{\textbf{Performance of different VLMs on LIBERO.}}
\begin{tabular}{lcccc|c}
\toprule
\textbf{VLM} & \textbf{Spatial} & \textbf{Object} & \textbf{Goal} & \textbf{Long} & \textbf{Average} \\
\midrule
Qwen2VL-2B~\citep{wang2024qwen2vlenhancingvisionlanguagemodels} & $80.4\%$ & $88.6\%$ & $83.8\%$ & $43.0\%$ & $73.9\%$ \\
InternVL2.5-1B~\citep{chen2025expandingperformanceboundariesopensource} & $64.6\%$ & $84.8\%$ & $78.4\%$ & $46.2\%$ & $68.5\%$ \\
Eagle2-1B~\citep{li2025eagle2buildingposttraining} & $73.4\%$ & $85.4\%$ & $90.8\%$ & $46.6\%$ & $\mathbf{74.0\%}$ \\
\bottomrule
\end{tabular}
\label{tab:singlevla-results}
\end{table}

Although Qwen2VL is widely regarded as robust, Eagle2-1B achieves comparable or slightly better results while using roughly half the parameters and providing significantly faster inference. We therefore select \textbf{Eagle2-1B} as the VLM backbone for SingleVLA.

\begin{table}[ht]
\centering
\caption{\textbf{Performance of pretrained SingleVLA on LIBERO.}}
\begin{tabular}{lcccc|c}
\toprule
\textbf{Method} & \textbf{Spatial} & \textbf{Object} & \textbf{Goal} & \textbf{Long} & \textbf{Average} \\
\midrule
SingleVLA (Eagle2-1B, no pretraining) & $73.4\%$ & $85.4\%$ & $90.8\%$ & $46.6\%$ & $74.0\%$ \\
SingleVLA (pretrained) & $\mathbf{92.4\%}$ & $\mathbf{94.5\%}$ & $\mathbf{93.5\%}$ & $\mathbf{63.7\%}$ & $\mathbf{86.0\%}$ \\
OpenVLA~\citep{kim2024openvla} & $84.7\%$ & $88.4\%$ & $79.2\%$ & $53.7\%$ & $76.5\%$ \\
Octo~\citep{octo_2023} & $78.9\%$ & $85.7\%$ & $84.6\%$ & $51.1\%$ & $75.1\%$ \\
\bottomrule
\end{tabular}
\label{tab:singlevla-results-finetuned}
\end{table}

After pretraining SingleVLA with Eagle2-1B, we fine-tune it on LIBERO to assess single-arm capability. As shown in Table~\ref{tab:singlevla-results-finetuned}, the pretrained SingleVLA substantially improves performance and even surpasses the 7B model OpenVLA, indicating that the learned single-arm policy is both effective and sufficiently strong to benefit the bimanual policy.

\section{Training Details}
\label{sec:training_details}

\begin{table}[ht]
\centering
\label{tab:baseline_hyperparameters}
\caption{\textbf{Training hyperparameters for baseline models.}}
\begin{tabular}{lccccc}
\toprule
\textbf{Method} & \textbf{\# of params} & \textbf{Learning rate} & \textbf{Lr scheduler} & \textbf{Batch size} & \textbf{Training steps} \\
\midrule
TwinVLA & 1.3B   & 1e-4   & cosine   & 8 & 100k \\
RDT-1B  & 1.2B   & 1e-4   & constant & 8 & 100k \\
DP      & 271M   & 2e-5   & cosine   & 8 & 100k \\
$\pi_0$ & 3.3B   & 2.5e-5 & cosine   & 8 & 100k \\
\bottomrule
\end{tabular}    
\end{table}

We use the official implementation of \textsc{RDT-1B}. Diffusion Policy and $\pi_0$ are evaluated via the public \textsc{LeRobot} release~\citep{cadene2024lerobot}, with two modifications for a fair comparison. First, the \textsc{LeRobot} evaluation script normalized images differently from training; we corrected this to match the training pipeline.

All models are fine-tuned with the same number of steps and batch size so that the total number of training samples is consistent across methods. For learning rates, we began with each model’s default and tuned within a similar compute budget. In practice, defaults worked well for DP and RDT-1B. For $\pi_0$, we observed better final returns by slowing the cosine decay; we therefore extended the LR schedule from 30k to 100k steps.

\section{TwinVLA Details}
\label{sec:implementation_details}

\begin{algorithm}
\caption{Joint Attention}
\label{alg:joint_attn_alg}
\begin{algorithmic}[1]
\Function{\color{jointAttnColor}JointAttention}{$\{Q_m\}, \{K_m\}, \{V_m\}, M$}
\State {$Q, K, V \leftarrow \text{Concatenate}(\{Q_m\},\{K_m\},\{V_m\})$ \Comment{Concatenate modality-specific Q, K, V}}
\State {$S \leftarrow \mathrm{Softmax}((QK^\top / \sqrt{d_k}) + M)$ \Comment{Apply causal joint mask M (\Cref{fig:joint_right})}}
\State {$S \leftarrow \text{\color{jointAttnColor}ApplyReweighting}(S)$ \Comment{Apply re-weighting (\Cref{alg:attention_reweighting_simple})}}
\State {$A \leftarrow S \cdot V$ \Comment{Calculate output A}}
\State \Return {$\{A_m\} \leftarrow \text{Split}(A)$ \Comment{Split output $A$ into modality-specific $A_m$}}
\EndFunction
\end{algorithmic}
\end{algorithm}

\subsection{Joint Attention}

The joint attention in TwinVLA is fundamentally almost identical to the implementation in the Mixture-of-Transformers (MoT)~\citep{liang2024mixtureoftransformerssparsescalablearchitecture}, but we applied attention-reweighting (\Cref{appendix:attn_re}). While MoT has transformers for text, image, and speech inputs, in TwinVLA, the inputs for the left and right arms correspond to these.

Furthermore, MoT requires an operation to group mixed inputs by modality and then restore their original order. However, this process is unnecessary in TwinVLA because the inputs are fed in a fixed sequence: left arm, then right arm. The detailed computation process is shown in~\Cref{alg:joint_attn_alg}.

\subsection{MoE integration}
\label{appendix:moe}

To enable sharing of the shared inputs between the two-arm models, we duplicated the entire VLM transformer. This necessitates different strategies for sharing the FFNs and the other components. This section details the strategy used for each component of the transformer.

\textbf{Feed-Forward Networks.} To share FFNs, we adopt the common approach of using a gating-based MoE. In standard MoE, multiple FFNs are included within a transformer, and a gating mechanism activates a subset for each input. In TwinVLA, the two VLMs act as distinct FFN experts.

Because shared inputs (e.g., ego-centric views or language prompts) may have asymmetric relevance for each arm, the gating mechanism learns how much each FFN should contribute to processing the shared input. This approach is widely used and has been shown to improve training stability and preserve information more effectively than simple averaging~\citep{shazeer2017outrageously}. We computed $w_\text{left}$ by applying a simple linear layer and softmax to the token embeddings.

\textbf{Other Components.} Beyond FFNs, elements such as layer normalization and projection layers also require integration. For these, we apply task arithmetic~\citep{10.5555/3692070.3694018}, merging the two VLMs via simple parameter averaging with weight $\lambda = 0.5$, elaborated~\Cref{alg:arith_alg}. This extends MoE-style computation to the full transformer architecture.

\begin{algorithm}
\caption{Integration of other components}
\label{alg:arith_alg}
\begin{algorithmic}[1]
\State Let $\text{Projection}_b$ be projection layer from each backbone $b \in \{\text{left}, \text{right}\}$.
\State Let $\text{LayerNorm}_b$ be layernorm from each backbone $b \in \{\text{left}, \text{right}\}$.
\Function{\color{projColor}Proj}{$X^m$}
\If{$m = \text{shared}$}
    \State {$F^m \leftarrow 0.5 \cdot (\text{Projection}_\text{left}(X^m) + \text{Projection}_\text{right}(X^m))$} \Comment{Task arithmetic}
\Else
    \State $F^m \leftarrow \text{Projection}_{m}(X^m)$ 
\EndIf
\State \Return {$F^m$}
\EndFunction
\State
\Function{\color{projColor}Norm}{$X^m$}
\If{$m = \text{shared}$}
    \State {$F^m \leftarrow 0.5 \cdot (\text{LayerNorm}_\text{left}(X^m) + \text{LayerNorm}_\text{right}(X^m))$} \Comment{Task arithmetic}
\Else
    \State $F^m \leftarrow \text{LayerNorm}_{m}(X^m)$ 
\EndIf
\State \Return {$F^m$}
\EndFunction
\end{algorithmic}
\end{algorithm}

\subsection{Attention Re-weighting}
\label{appendix:attn_re}

\begin{figure}[h]
    \centering
    \includegraphics[width=\textwidth]{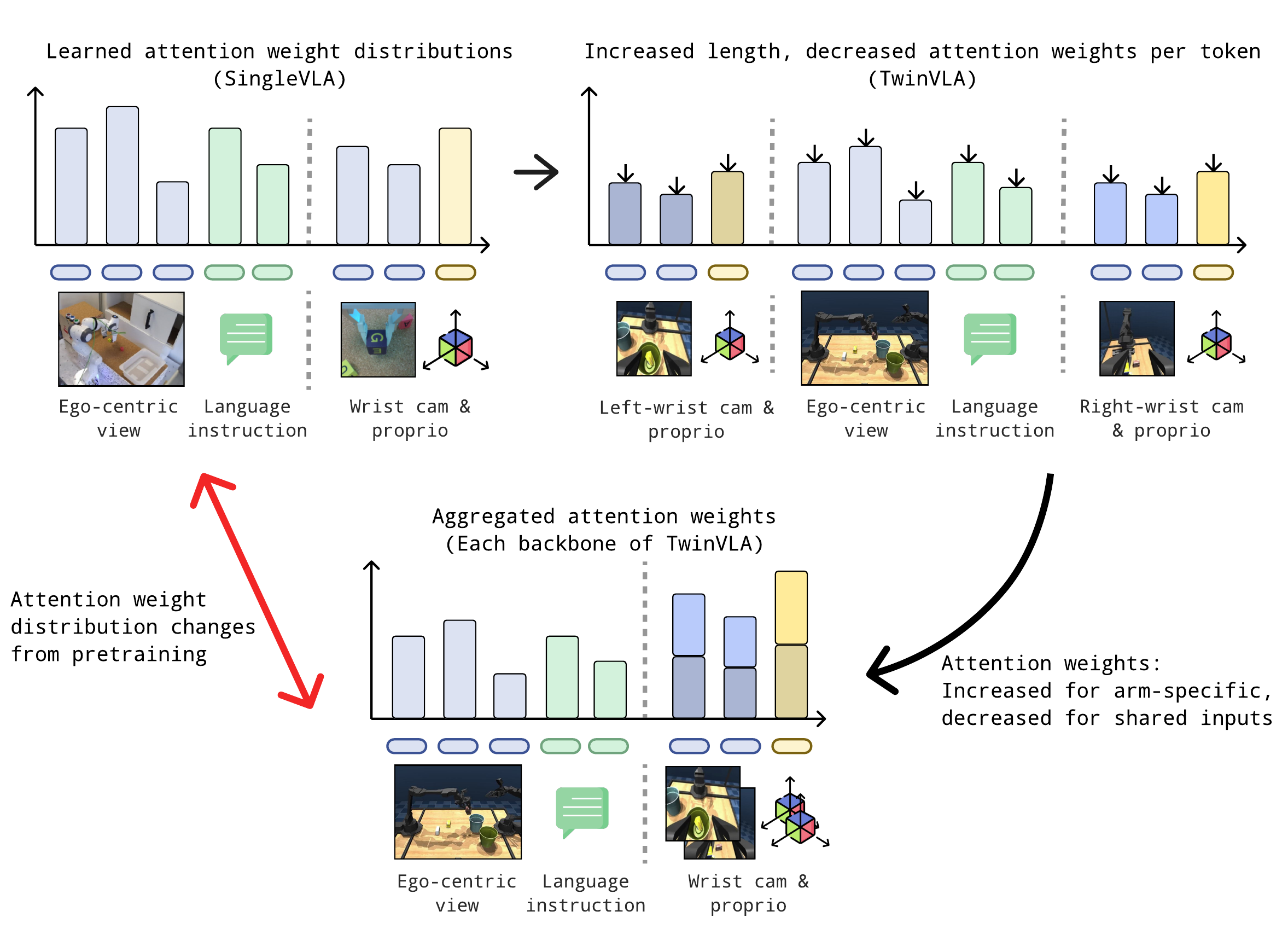}
    \caption{Due to the increased token length and softmax normalization, each VLM of TwinVLA refers to arm-specific inputs more than during pretraining, requiring the model to adapt.}
    \label{fig:attn_re_1}
\end{figure}

\begin{algorithm}[h!]
\caption{Attention Re-weighting}
\label{alg:attention_reweighting_simple}
\begin{algorithmic}[2]
\Function{\color{jointAttnColor}ApplyReweighting}{$\mathbf{A}$, $\alpha=2$}
    \State Create mask $\mathbf{M_r} = (m \not= \text{shared})$
    \Comment{Create a mask for arm-specific inputs}
    
    \State $\mathbf{A_{reweighted}} \gets \mathbf{A} \odot (\mathbf{M_r} + \alpha \cdot \neg \mathbf{M_r})$
    \Comment{Apply scaling to attention weights using the mask}
    
    \State $\mathbf{A_{reweighted}} \gets \text{Normalize}(\mathbf{A_{reweighted}})$
    \Comment{Normalize the new weights}
    
    \State \Return $\mathbf{A} + (\mathbf{A_{reweighted}} - \mathbf{A})$
    \Comment{Return weights as a residual update for gradient flow}
\EndFunction
\end{algorithmic}
\end{algorithm}

Attention re-weighting is a technique we employ to improve the efficiency of adapting a pretrained SingleVLA into a bimanual TwinVLA. Constructing TwinVLA involves adding a second set of arm-specific modality tokens. During operation, input tokens are processed by their corresponding arm’s VLM backbone, pass through a joint attention layer, and then flow back to the individual VLMs. However, the softmax normalization within this joint attention layer presents a challenge. Although the total sequence length doubles, the number of tokens for shared inputs remains unchanged. Consequently, the proportion of attention allocated to these shared inputs is significantly diluted compared to the pretraining phase, creating a distribution shift for each VLM backbone’s inputs, as illustrated in \Cref{fig:attn_re_1}.

\begin{figure}[h]
    \centering
    \includegraphics[width=\textwidth]{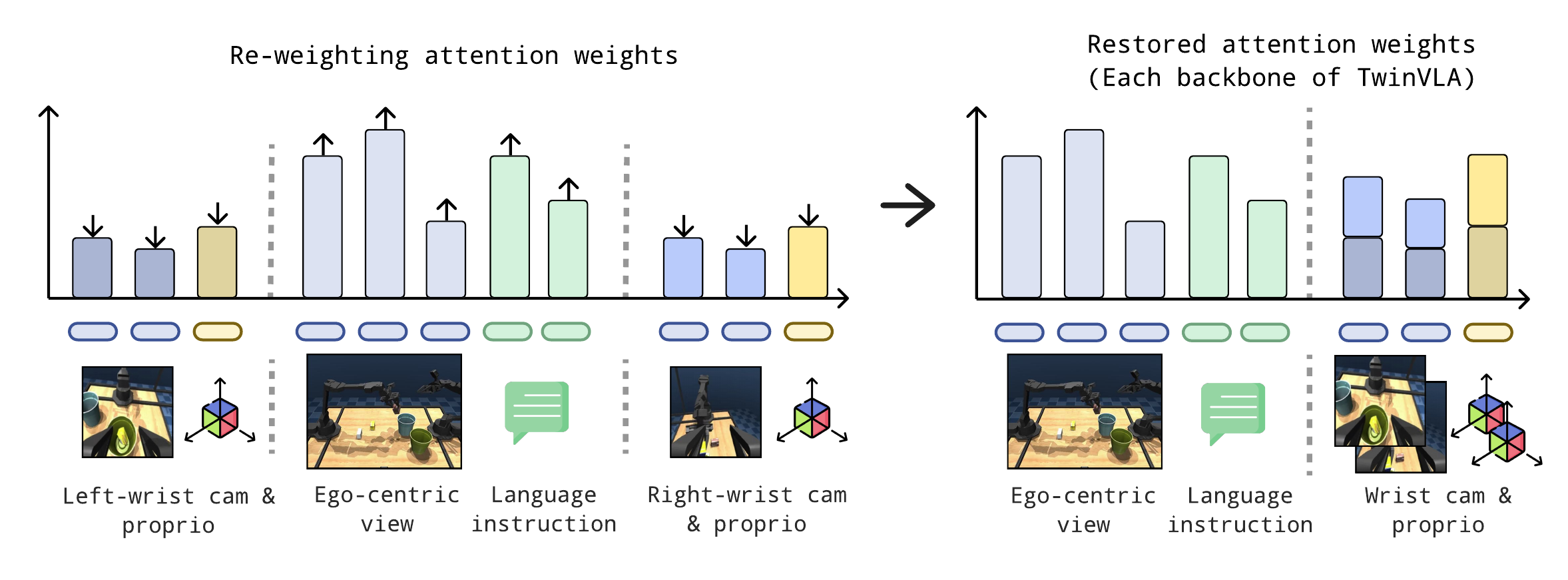}
    \caption{By re-weighting the attention weights, we can make each VLM refer to each modality identically to its pretraining stage, resulting in no adaptation and a lower initial loss.}
    \label{fig:attn_re_2}
\end{figure}

This discrepancy requires greater adaptation effort for TwinVLA during fine-tuning on bimanual tasks. To address this, we introduce a simple re-weighting trick immediately after the attention scores are calculated. Specifically, we double the attention weights corresponding to the shared modality tokens and then re-normalize all weights to sum to one. This adjustment effectively restores the proportional attention each VLM backbone assigns to the shared inputs, aligning it with the pretraining conditions (see~\Cref{fig:attn_re_2}). Applying this method reduced the initial fine-tuning loss by approximately 40\%. While TwinVLA could learn bimanual manipulation without this technique, the required adaptation period would be substantially longer. This simple trick makes the process significantly more efficient and faster. We illustrate our implementation with simple pseudocode in~\Cref{alg:attention_reweighting_simple}.

\section{Real-World Robot Experiment Details}
\label{sec:realworld_details}

\subsection{Task details}

\begin{figure}[h]
  \centering
  \includegraphics[width=1.0\linewidth]{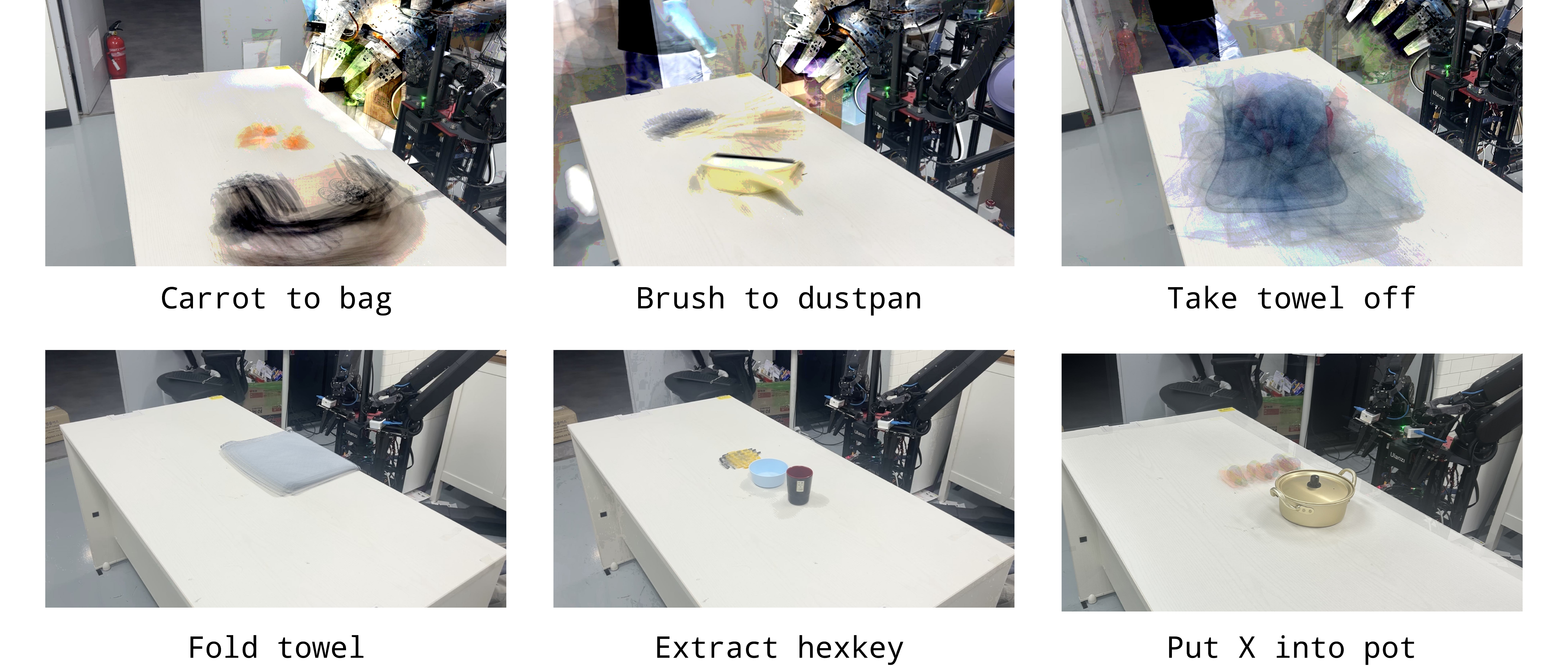}
  \caption{Initial distribution of each tasks in real-world.}
  \label{fig:real-distribution}
\end{figure}

To illustrate the diversity of initial configurations in our dataset, \Cref{fig:real-distribution} shows an overlay of the first frames from all 50 demonstrations. For each demonstration, the position and orientation of the objects were randomized, resulting in a unique starting setup.

\begin{figure}[h]
  \centering
  \includegraphics[width=0.6\linewidth]{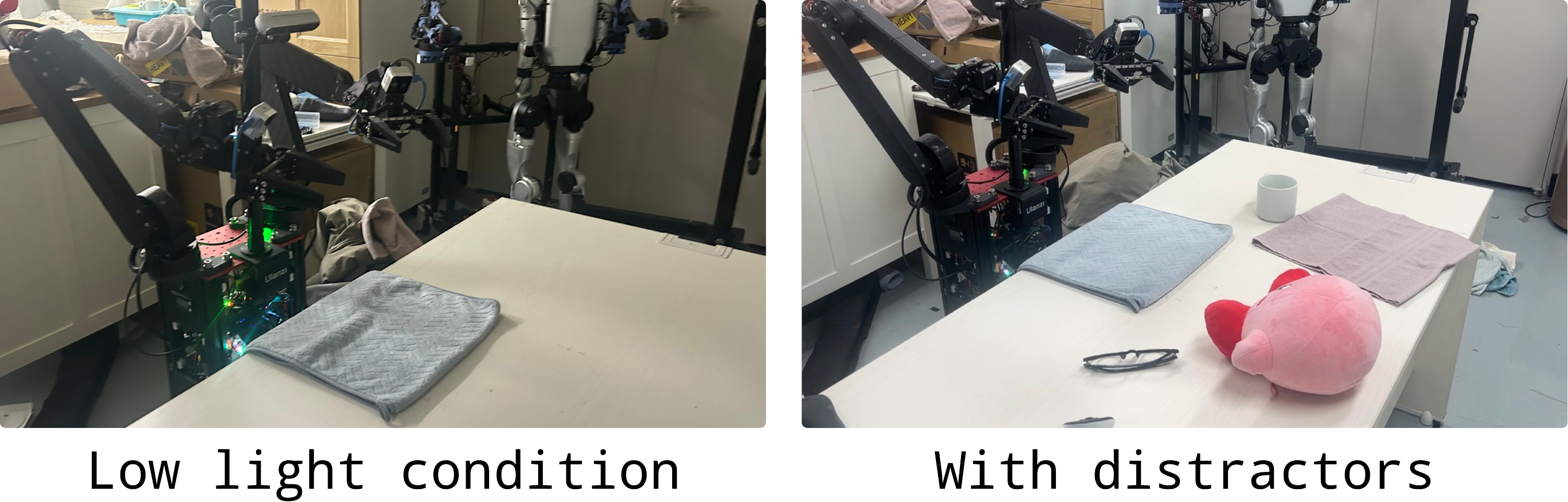}
  \caption{Challenging scene of \texttt{Fold towel} task.}
  \label{fig:fold-rob}
\end{figure}

Furthermore, to evaluate policy robustness in the real world, we tested the \texttt{Fold towel} task under more challenging conditions, such as reduced lighting and the presence of distractors. These scenarios are visualized in~\Cref{fig:fold-rob}.

\subsection{Quantitative Results}

\begin{table}[h]
\centering
\caption{\textbf{Success rates for each model across all subtasks.} The best overall performance is highlighted in bold. As $\pi_0$ is included as an upper-bound, as this is excluded from this direct comparison.}
\begin{tabular}{llccccc}
\toprule
\textbf{Task} & \textbf{Subtask} & \textbf{DP} & \textbf{TwinVLA} & \textbf{RDT-1B} & $\boldsymbol{\pi_0}$ \\
\midrule
\multirow{3}{*}{\texttt{Fold towel}}
  & \texttt{First fold} & $0.00$ & $1.00$ & $0.90$ & $1.00$\\
  & \texttt{Rotate}     & $0.00$ & $1.00$ & $0.85$ & $1.00$\\
  & \texttt{Second fold}      & $0.00$ & $\textbf{0.90}$ & $0.45$ & $\textbf{0.90}$\\
\midrule
\multirow{3}{*}{\texttt{Extract hexkey}}
  & \texttt{Pick up} & $0.60$ & $0.90$ & $0.90$ & $1.00$\\
  & \texttt{Extract}     & $0.35$ & $0.80$ & $0.55$ & $0.90$\\
  & \texttt{Put into bowl}      & $0.30$ & $\textbf{0.80}$ & $0.45$ & $\textbf{0.80}$\\
\midrule
\multirow{3}{*}{\texttt{Carrot to bag}}
  & \texttt{Pick up carrot} & $0.50$ & $1.00$ & $0.75$ & $0.85$\\
  & \texttt{Put carrot}     & $0.20$ & $0.70$ & $0.40$ & $0.65$\\
  & \texttt{Close bag}      & $0.15$ & $0.60$ & $0.35$ & $\textbf{0.65}$\\
\midrule
\multirow{3}{*}{\texttt{Brush to dustpan}}
  & \texttt{Move the brush}   & $0.70$ & $1.00$ & $1.00$ & $1.00$\\
  & \texttt{Pick up the brush} & $0.65$ & $1.00$ & $1.00$ & $1.00$\\
  & \texttt{Put onto dustpan} & $0.35$ & $\textbf{0.80}$ & 0.40 & $\textbf{0.80}$\\
\midrule
\multirow{3}{*}{\texttt{Take towel off}}
  & \texttt{Dragging}         & $0.40$ & $0.90$ & $0.80$ & $0.95$ \\
  & \texttt{Half off}         & $0.35$ & $0.70$ & $0.70$ & $0.85$ \\
  & \texttt{Entirely off}     & $0.20$ & $0.45$ & $0.60$ & \textbf{0.65}\\
\bottomrule
\end{tabular}
\label{tab:carrot_to_bag}
\end{table}

We provide the quantitative results on real-world experiments in subtask-level detail in~\Cref{tab:carrot_to_bag}. The results reveal the main bottleneck in each long-horizon task. First, for the two tasks, \texttt{Fold towel} and \texttt{Extract hexkey}, requiring tightly coupled bimanual coordination, the phase where both arms meet to execute the action appears to be critical. The \texttt{Carrot to bag} task is challenging when inserting the carrot, which requires precisely opening the bag. The \texttt{Brush to dustpan} task's bottleneck is the high-precision insertion of the brush into the dustpan. Lastly, in \texttt{Take towel off}, the final unfolding is difficult---unlike the simple initial steps---as it requires a successful switch between the arms. In the next subsection, we show qualitative results from these specific bottleneck phases.

\subsection{Qualitative Results}
\begin{figure}[h]
  \centering
  \includegraphics[width=1.0\linewidth]{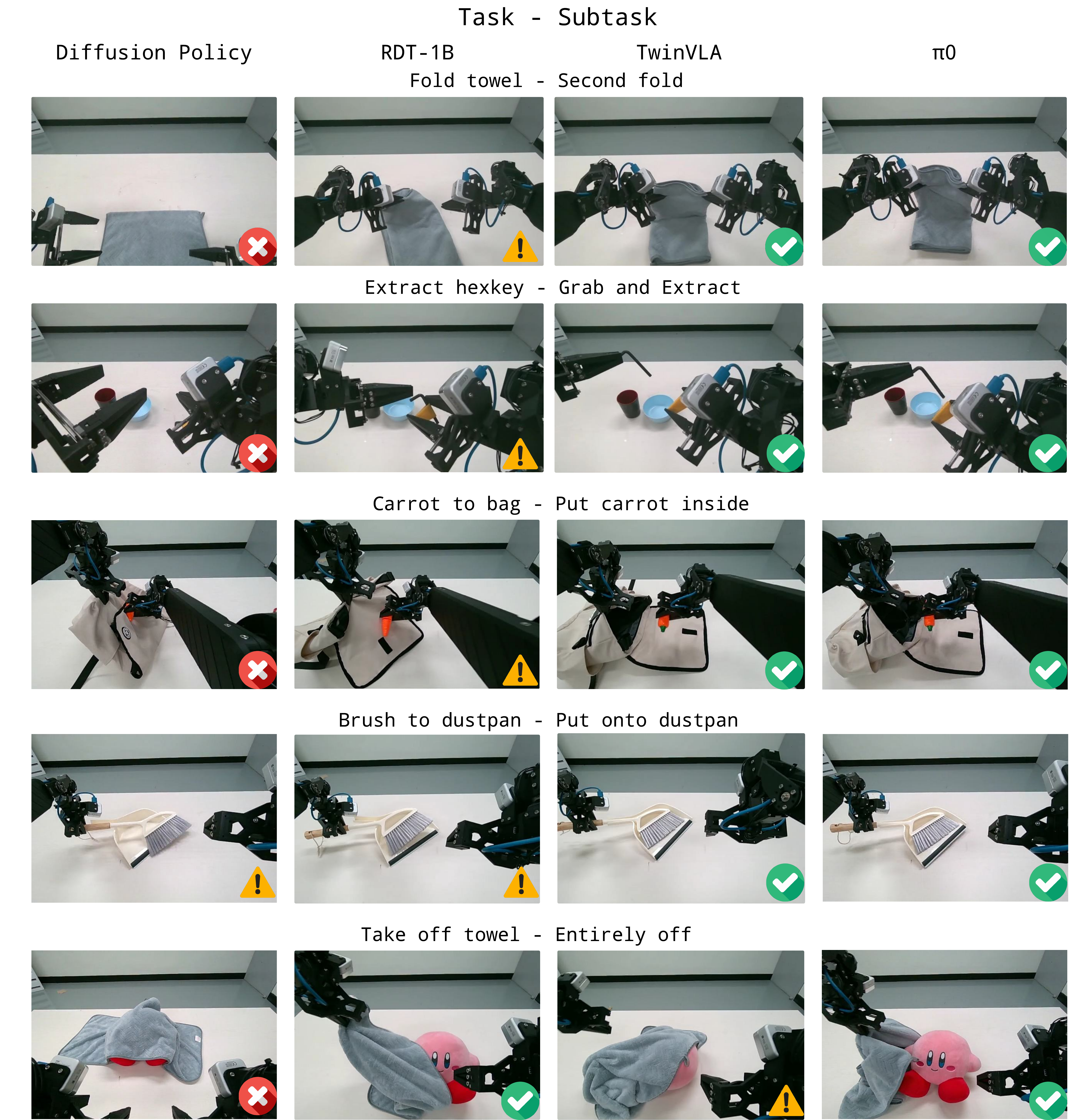}
  \caption{\textbf{Qualitative visualization of real world experiments.}}
  \label{fig:Quali}
\end{figure}

\Cref{fig:Quali} presents qualitative results highlighting challenging situations for each task. A check mark was used when the model succeeded with a probability above 0.5, an X mark for probabilities below 0.3, and an exclamation mark icon for intermediate cases.

\begin{itemize}[leftmargin=2em,topsep=0pt,itemsep=0pt,parsep=3pt]
    \item \textbf{\texttt{Carrot to bag.}} $\pi_0$ showed the highest success rate, followed by TwinVLA, RDT, and DP. DP failed to interact meaningfully with the bag, especially struggling to grasp the cover properly. RDT failed to complete the task successfully, primarily due to its inability to accurately localize and grasp the bag's opening.
    \item \textbf{\texttt{Brush to dustpan.}} DP struggled either to grasp the brush itself or to successfully insert it. Interestingly, the RDT managed to grasp the brush well but lacked precision during the insertion. In this task, TwinVLA and $\pi_0$ demonstrated the same success rate.
    \item \textbf{\texttt{Take towel off.}} DP mostly failed to pull the doll from a distant position toward the center, while the other models succeeded in pulling it to the center but showed differences in towel removal. Both RDT and $\pi_0$ tended to successfully remove one side of the towel and then easily remove the other side as well. In contrast, TwinVLA struggled with removing the remaining part and repeated the same action. This is likely because the longer action chunk length of RDT and $\pi_0$ helped them overcome the multimodality challenge.
    \item \textbf{\texttt{Fold towel.}} $\pi_0$ and TwinVLA successfully completed the task. RDT also generally performed well, but occasionally failed to fully rotate the towel by 90 degrees, which caused downstream failures. DP experienced substantial difficulty with the fold-towel task and ultimately failed to solve it.
    \item \textbf{\texttt{Extract hexkey.}} $\pi_0$ and TwinVLA generally solved the task reliably. RDT performed the subtask of lifting the hexkey case well but often failed during extraction due to insufficient precision in grasping the hexkey once the case was lifted. DP failed both to reliably pick up the hexkey case and to extract the hexkey itself.
\end{itemize}

\subsection{Robot Hardware Spec}

We conduct our real-world experiments using a custom-built robot named Anubis. The platform features a teleoperation system inspired by the Mobile ALOHA setup~\citep{fu2024mobile}. Each arm has 6 DoF and is equipped with a parallel gripper and a wrist-mounted camera. At the center of the robot, an Intel RealSense camera is mounted on a height-adjustable mechanism, serving as the ego-centric view camera. Details are described in~\Cref{tab:anubis_hardware}. Anubis is equipped with a 3-wheel omni-directional base that supports planar locomotion; however, in this work, the mobility feature is not utilized.

\begin{table}[H]
  \centering
  \begin{minipage}[b]{0.65\textwidth}
    \centering
    \footnotesize
    \setlength{\tabcolsep}{4pt}
    \vspace{1em}
    \caption{\textbf{Anubis Robot Hardware Specifications.}}
    \begin{tabularx}{\linewidth}{@{}lX@{}}
      \toprule
      \textbf{Component}    & \textbf{Specification} \\
      \midrule
      Base Type             & 3-wheel omni-directional chassis \\
      Mobility DOF          & 3 (X, Y, Yaw) \\
      Arm DOF               & 2 × (6 DOF + gripper) = 14 \\
      Total Action Space    & 17 DOF \\
      Wrist Cameras         & Intel RealSense D405 \\
      Gripper               & Parallel transparent gripper (hole design, ALOHA-style) \\
      Power System          & 3 × Greenworks 40V 5.0Ah batteries (PC, wheels \& leader/follower) \\
      Frame                 & 3D-printed custom components \\
      \bottomrule
    \end{tabularx}
    \label{tab:anubis_hardware}
  \end{minipage}
  \hfill
  \begin{minipage}[b]{0.30\textwidth}
    \centering
    \includegraphics[width=0.6\linewidth]{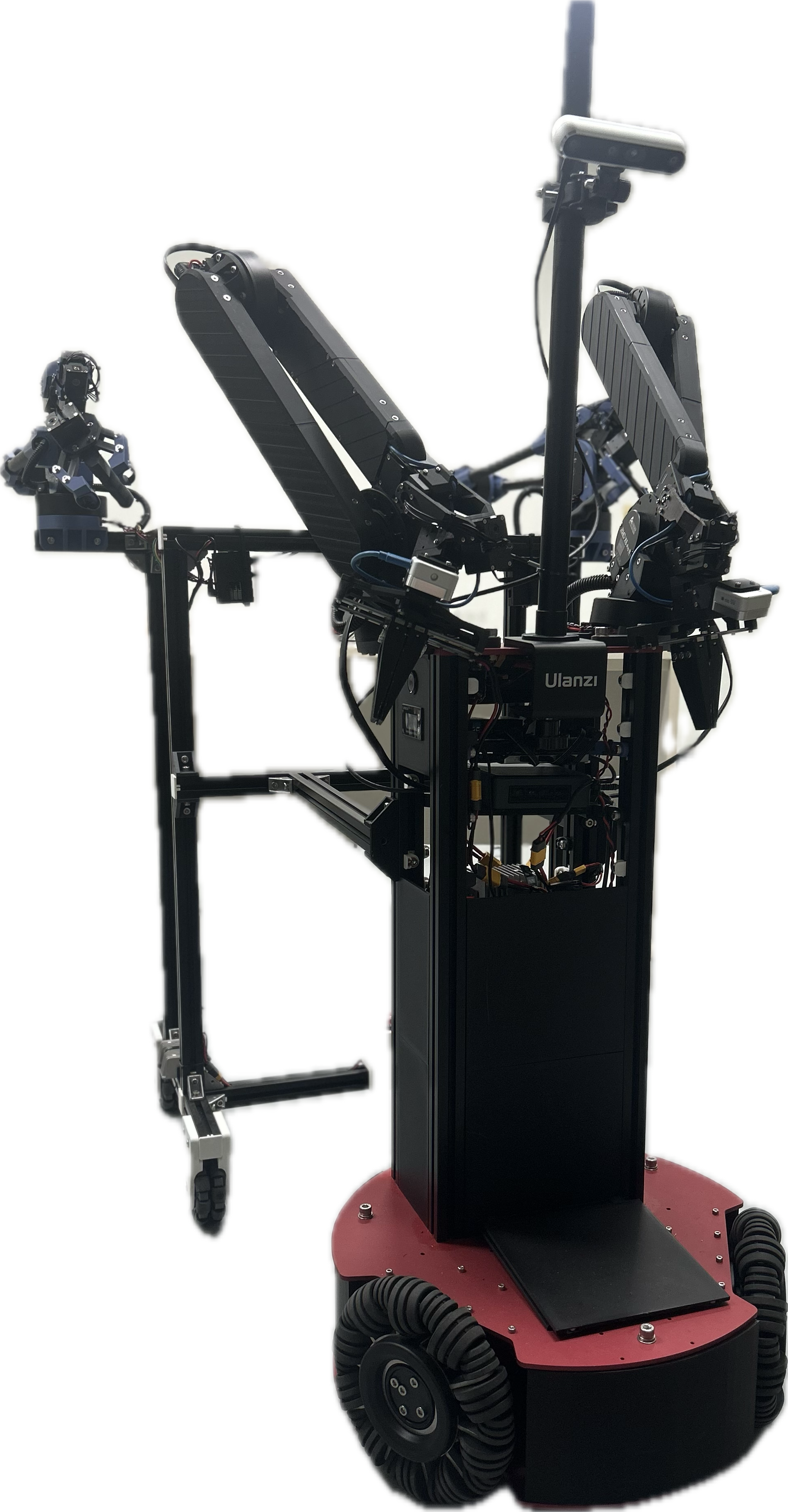} 
    \vspace{1em}
    \captionof{figure}{\textbf{The Anubis robot.}}
    \label{fig:anubis}
  \end{minipage}
\end{table}

\section{Simulation Experiment Details}
\label{fig:sim_exp_details}

\begin{figure}[h]
  \centering
  \includegraphics[width=1.0\linewidth]{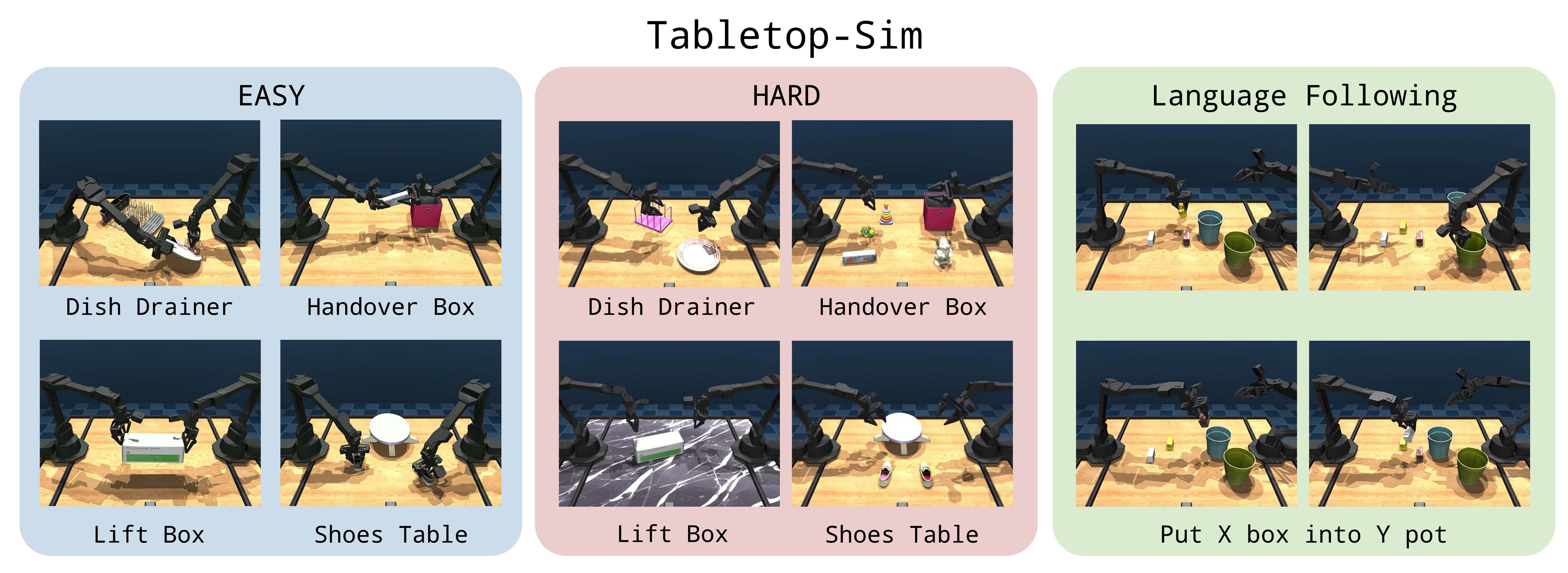}
  \caption{\textbf{Task list of Tabletop-Sim.}}
  \label{fig:tabletop-sim}
\end{figure}

\subsection{Tabletop-Sim}

To test bimanual policies in simulation, we developed Tabletop-Sim, a new benchmark specifically engineered to evaluate dexterous manipulation skills, in contrast to other benchmarks~\citep{mu2025robotwindualarmrobotbenchmark} that primarily focus on task diversity. The code are publicly available at \url{https://github.com/jellyho/Tabletop-Sim}. The benchmark comprises four single-task environments and one multi-task setup.  Our task selection was guided by the taxonomy in DexMimicGen~\citep{11127809}, which categorizes bimanual tasks into: (1) parallel (two arms are doing separate tasks simultaneously), (2) coordinated (two arms are closely working together), and (3) sequential (one arm completes the task, and the other arm takes over) interactions. Using a custom controller similar to GELLO~\citep{wu2024gellogenerallowcostintuitive}, we collected 50 demonstrations for each single-task and 60 for the multi-task environment.

The multi-task setup is a language-following task requiring the policy to place a specific box (out of three) into a designated pot (out of two) based on a language instruction. This task is designed to rigorously assess a model's instruction-following capabilities, as Vision-Language-Action (VLA) models often disregard instructions after fine-tuning.

Furthermore, to evaluate policy robustness, we established two difficulty settings for the four single-tasks. The original tasks are designated as the Easy setting, while a Hard variant for each task incorporates challenging variations such as different textures, object models, and the presence of distractor objects. \Cref{fig:tabletop-sim} presents snapshots of each task.

\subsection{Quantitative Results}

This section describes the detailed results for the simulation tasks. The results for \textbf{Tabletop-Sim} are listed in \Cref{tab:aloha-sim-elegant}, while the results for the \textbf{RoboTwin 2.0} benchmark are in \Cref{tab:bimanual-tasks-final}. For RoboTwin, the results for other baselines were referenced from the official benchmark results.

Although $\pi_\textbf{0}$ achieves the highest overall performance, this result is unsurprising considering its larger model size and pretraining dataset. Meanwhile, \textbf{TwinVLA} demonstrates consistently superior performance compared to \textbf{RDT-1B}, a model of a similar scale.

\begin{table}[h]
\centering
\renewcommand{\arraystretch}{1.2}
\setlength{\tabcolsep}{4pt}

\caption{\textbf{Performance comparison on the Tabletop-Sim benchmark.}}
\label{tab:aloha-sim-elegant}
\begin{tabular}{@{} l cccccccc >{\centering\arraybackslash}p{2.5cm} @{}}
\toprule
& \multicolumn{8}{c}{Tabletop-Sim} & \\
\cmidrule(r){2-9}
& \multicolumn{2}{c}{\texttt{Dish drainer}} & \multicolumn{2}{c}{\texttt{Handover box}} & \multicolumn{2}{c}{\texttt{Lift box}} & \multicolumn{2}{c}{\texttt{Shoes table}} & \multirow{2}{*}{\parbox{2.5cm}{\centering \texttt{Put X cube in to Y pot}}} \\
\cmidrule(lr){2-3} \cmidrule(lr){4-5} \cmidrule(lr){6-7} \cmidrule(lr){8-9}
Model & Easy & Hard & Easy & Hard & Easy & Hard & Easy & Hard & \\
\midrule
DP & $0.686$ & $0.590$ & $0.180$ & $0.086$ & $0.100$ & $0.006$ & $0.028$ & $0.260$ & - \\
RDT-1B & $0.810$ & $0.780$ & $0.694$ & $0.508$ & $0.300$ & $0.076$ & $0.660$ & $0.192$ & $0.555$ \\
TwinVLA & $\textbf{0.954}$ & $\textbf{0.836}$ & $0.780$ & $\textbf{0.530}$ & $0.452$ & $0.044$ & $\textbf{0.848}$ & $0.306$ & $\textbf{0.806}$ \\
PI-0 & $0.774$ & $0.520$ & $\textbf{0.788}$ & $0.444$ & $\textbf{0.512}$ & $\textbf{0.136}$ & $0.824$ & $\textbf{0.660}$ & $0.792$ \\
\bottomrule
\end{tabular}
\end{table}

\begin{table}[h]
\centering
\caption{\textbf{Success rates of TwinVLA for $50$ bimanual tasks in RoboTwin 2.0.}}
\label{tab:bimanual-tasks-final}
\begin{tabular}{@{} l c c c l c c @{}}
\toprule
\texttt{Task Name} & Easy & Hard & & \texttt{Task Name} & Easy & Hard \\
\midrule
\texttt{adjust bottle} & $0.97$ & $0.35$ & & \texttt{place can basket} & $0.40$ & $0.00$ \\
\texttt{beat block hammer} & $0.77$ & $0.10$ & & \texttt{place cans plasticbox} & $0.47$ & $0.08$ \\
\texttt{blocks ranking rgb} & $0.58$ & $0.00$ & & \texttt{place container plate} & $0.77$ & $0.04$ \\
\texttt{blocks ranking size} & $0.03$ & $0.00$ & & \texttt{place dual shoes} & $0.18$ & $0.03$ \\
\texttt{click alarmclock} & $0.33$ & $0.01$ & & \texttt{place empty cup} & $0.50$ & $0.01$ \\
\texttt{click bell} & $0.58$ & $0.13$ & & \texttt{place fan} & $0.34$ & $0.00$ \\
\texttt{dump bin bigbin} & $0.80$ & $0.34$ & & \texttt{place mouse pad} & $0.50$ & $0.00$ \\
\texttt{grab roller} & $0.96$ & $0.22$ & & \texttt{place object basket} & $0.48$ & $0.03$ \\
\texttt{handover block} & $0.17$ & $0.00$ & & \texttt{place object scale} & $0.06$ & $0.00$ \\
\texttt{handover mic} & $0.84$ & $0.02$ & & \texttt{place object stand} & $0.20$ & $0.02$ \\
\texttt{hanging mug} & $0.10$ & $0.05$ & & \texttt{place phone stand} & $0.34$ & $0.02$ \\
\texttt{lift pot} & $0.87$ & $0.07$ & & \texttt{place shoe} & $0.48$ & $0.04$ \\
\texttt{move can pot} & $0.45$ & $0.05$ & & \texttt{press stapler} & $0.62$ & $0.26$ \\
\texttt{move pillbottle pad} & $0.32$ & $0.02$ & & \texttt{put bottles dustbin} & $0.08$ & $0.04$ \\
\texttt{move playingcard away} & $0.61$ & $0.35$ & & \texttt{put object cabinet} & $0.39$ & $0.16$ \\
\texttt{move stapler pad} & $0.11$ & $0.00$ & & \texttt{rotate qrcode} & $0.54$ & $0.03$\\
\texttt{open laptop} & $0.80$ & $0.17$ & & \texttt{scan object} & $0.11$ & $0.04$\\
\texttt{open microwave} & $0.03$ & $0.01$ & & \texttt{shake bottle horizontally} & $0.96$ & $0.55$\\
\texttt{pick diverse bottles} & $0.16$ & $0.08$ & & \texttt{shake bottle} & $0.93$ & $0.58$\\
\texttt{pick dual bottles} & $0.18$ & $0.12$ & & \texttt{stack blocks three} & $0.00$ & $0.00$\\
\texttt{place a2b left} & $0.27$ & $0.05$ & & \texttt{stack blocks two} & $0.26$ & $0.00$ \\
\texttt{place a2b right} & $0.15$ & $0.01$ & & \texttt{stack bowls three} & $0.77$ & $0.15$ \\
\texttt{place bread basket} & $0.11$ & $0.03$ & & \texttt{stack bowls two} & $0.84$ & $0.11$ \\
\texttt{place bread skillet} & $0.20$ & $0.01$ & & \texttt{stamp seal} & $0.16$ & $0.01$ \\
\texttt{place burger fries} & $0.67$ & $0.13$ & & \texttt{turn switch} & $0.25$ & $0.15$ \\
\midrule
\textbf{Average} &  &  \\
\midrule
Diffusion Policy & $0.280$ & $0.006$ \\
RDT-1B & $0.345$ & $0.137$ \\
TwinVLA & $0.420$ & $0.089$ \\
$\pi_0$ & $0.464$ & $0.163$ \\
\bottomrule
\end{tabular}
\end{table}

\end{document}

%% file: math_commands.tex
\usepackage{amsmath,amsfonts,bm}

\def\eqref#1{equation~\ref{#1}}

\def\1{\bm{1}}

\DeclareMathAlphabet{\mathsfit}{\encodingdefault}{\sfdefault}{m}{sl}
\SetMathAlphabet{\mathsfit}{bold}{\encodingdefault}{\sfdefault}{bx}{n}



%% file: iclr2026_conference.bib
@string{CVPR = "IEEE/CVF Conference on Computer Vision and Pattern Recognition"}

@string{ICCV = "IEEE/CVF International Conference on Computer Vision"}

@string{NeurIPS = "Advances in Neural Information Processing Systems"}

@string{ICLR = "International Conference on Learning Representations"}

@string{ICML = "International Conference on Machine Learning"}

@string{CPAL = "Conference on Parsimony and Learning"}

@string{CoRL = "Conference on Robot Learning"}

@string{RSS = "Robotics: Science and Systems"}

@string{ICRA = "IEEE International Conference on Robotics and Automation"}

@string{IROS = "IEEE/RSJ International Conference on Intelligent Robots and Systems"}

@string{Humanoids = "IEEE-RAS International Conference on Humanoid Robots"}

@string{IJRR = "The International Journal of Robotics Research"}

@string{RAL = "IEEE Robotics and Automation Letters"}

@string{TC = "IEEE Transactions on Computers"}

@string{SMC = "IEEE International Conference on Systems, Man, and Cybernetics"}

@article{chi2023diffusion,
	author = {Cheng Chi and Zhenjia Xu and Siyuan Feng and Eric Cousineau and Yilun Du and Benjamin Burchfiel and Russ Tedrake and Shuran Song},
	title ={Diffusion Policy: Visuomotor Policy Learning via Action Diffusion},
	journal = IJRR,
	year = {2024},
}

@inproceedings{zhao2023learning,
  title={Learning fine-grained bimanual manipulation with low-cost hardware},
  author={Zhao, Tony Z and Kumar, Vikash and Levine, Sergey and Finn, Chelsea},
  booktitle=RSS,
  year={2023}
}

@inproceedings{lee2024interact,
  title={Inter{ACT}: Inter-dependency Aware Action Chunking with Hierarchical Attention Transformers for Bimanual Manipulation},
  author={Andrew Choong-Won Lee and Ian Chuang and Ling-Yuan Chen and Iman Soltani},
  booktitle=CoRL,
  year={2024},
  url={https://openreview.net/forum?id=lKGRPJFPCM}
}

@inproceedings{black2024pi_0,
  title={$\pi_0$: A Vision-Language-Action Flow Model for General Robot Control},
  author={Black, Kevin and Brown, Noah and Driess, Danny and Esmail, Adnan and Equi, Michael and Finn, Chelsea and Fusai, Niccolo and Groom, Lachy and Hausman, Karol and Ichter, Brian and others},
  booktitle=RSS,
  year={2024}
}

@inproceedings{khazatsky2024droid,
  title={DROID: A Large-Scale In-The-Wild Robot Manipulation Dataset},
  author={Alexander Khazatsky and Karl Pertsch and Suraj Nair and Ashwin Balakrishna and Sudeep Dasari and Siddharth Karamcheti and Soroush Nasiriany and Mohan Kumar Srirama and Lawrence Yunliang Chen and Kirsty Ellis and Peter David Fagan and Joey Hejna and Masha Itkina and Marion Lepert and Yecheng Jason Ma and Patrick Tree Miller and Jimmy Wu and Suneel Belkhale and Shivin Dass and Huy Ha and Arhan Jain and Abraham Lee and Youngwoon Lee and Marius Memmel and Sungjae Park and Ilija Radosavovic and Kaiyuan Wang and Albert Zhan and Kevin Black and Cheng Chi and Kyle Beltran Hatch and Shan Lin and Jingpei Lu and Jean Mercat and Abdul Rehman and Pannag R Sanketi and Archit Sharma and Cody Simpson and Quan Vuong and Homer Rich Walke and Blake Wulfe and Ted Xiao and Jonathan Heewon Yang and Arefeh Yavary and Tony Z. Zhao and Christopher Agia and Rohan Baijal and Mateo Guaman Castro and Daphne Chen and Qiuyu Chen and Trinity Chung and Jaimyn Drake and Ethan Paul Foster and Jensen Gao and David Antonio Herrera and Minho Heo and Kyle Hsu and Jiaheng Hu and Donovon Jackson and Charlotte Le and Yunshuang Li and Kevin Lin and Roy Lin and Zehan Ma and Abhiram Maddukuri and Suvir Mirchandani and Daniel Morton and Tony Nguyen and Abigail O'Neill and Rosario Scalise and Derick Seale and Victor Son and Stephen Tian and Emi Tran and Andrew E. Wang and Yilin Wu and Annie Xie and Jingyun Yang and Patrick Yin and Yunchu Zhang and Osbert Bastani and Glen Berseth and Jeannette Bohg and Ken Goldberg and Abhinav Gupta and Abhishek Gupta and Dinesh Jayaraman and Joseph J Lim and Jitendra Malik and Roberto Martín-Martín and Subramanian Ramamoorthy and Dorsa Sadigh and Shuran Song and Jiajun Wu and Michael C. Yip and Yuke Zhu and Thomas Kollar and Sergey Levine and Chelsea Finn},
  booktitle=RSS,
  year={2024}
}

@inproceedings{kim2024openvla,
  title={Open{VLA}: An Open-Source Vision-Language-Action Model},
  author={Moo Jin Kim and Karl Pertsch and Siddharth Karamcheti and Ted Xiao and Ashwin Balakrishna and Suraj Nair and Rafael Rafailov and Ethan P Foster and Pannag R Sanketi and Quan Vuong and Thomas Kollar and Benjamin Burchfiel and Russ Tedrake and Dorsa Sadigh and Sergey Levine and Percy Liang and Chelsea Finn},
  booktitle=CoRL,
  year={2024},
}

@inproceedings{oxe2024,
  title={Open {X-E}mbodiment: Robotic Learning Datasets and {RT-X} Models},
  author={{Open X-Embodiment Collaboration} and Abby O'Neill and Abdul Rehman and Abhinav Gupta and Abhiram Maddukuri and Abhishek Gupta and Abhishek Padalkar and Abraham Lee and Acorn Pooley and Agrim Gupta and Ajay Mandlekar and Ajinkya Jain and Albert Tung and Alex Bewley and Alex Herzog and Alex Irpan and Alexander Khazatsky and Anant Rai and Anchit Gupta and Andrew Wang and Andrey Kolobov and Anikait Singh and Animesh Garg and Aniruddha Kembhavi and Annie Xie and Anthony Brohan and Antonin Raffin and Archit Sharma and Arefeh Yavary and Arhan Jain and Ashwin Balakrishna and Ayzaan Wahid and Ben Burgess-Limerick and Beomjoon Kim and Bernhard Schölkopf and Blake Wulfe and Brian Ichter and Cewu Lu and Charles Xu and Charlotte Le and Chelsea Finn and Chen Wang and Chenfeng Xu and Cheng Chi and Chenguang Huang and Christine Chan and Christopher Agia and Chuer Pan and Chuyuan Fu and Coline Devin and Danfei Xu and Daniel Morton and Danny Driess and Daphne Chen and Deepak Pathak and Dhruv Shah and Dieter Büchler and Dinesh Jayaraman and Dmitry Kalashnikov and Dorsa Sadigh and Edward Johns and Ethan Foster and Fangchen Liu and Federico Ceola and Fei Xia and Feiyu Zhao and Felipe Vieira Frujeri and Freek Stulp and Gaoyue Zhou and Gaurav S. Sukhatme and Gautam Salhotra and Ge Yan and Gilbert Feng and Giulio Schiavi and Glen Berseth and Gregory Kahn and Guangwen Yang and Guanzhi Wang and Hao Su and Hao-Shu Fang and Haochen Shi and Henghui Bao and Heni Ben Amor and Henrik I Christensen and Hiroki Furuta and Homanga Bharadhwaj and Homer Walke and Hongjie Fang and Huy Ha and Igor Mordatch and Ilija Radosavovic and Isabel Leal and Jacky Liang and Jad Abou-Chakra and Jaehyung Kim and Jaimyn Drake and Jan Peters and Jan Schneider and Jasmine Hsu and Jay Vakil and Jeannette Bohg and Jeffrey Bingham and Jeffrey Wu and Jensen Gao and Jiaheng Hu and Jiajun Wu and Jialin Wu and Jiankai Sun and Jianlan Luo and Jiayuan Gu and Jie Tan and Jihoon Oh and Jimmy Wu and Jingpei Lu and Jingyun Yang and Jitendra Malik and João Silvério and Joey Hejna and Jonathan Booher and Jonathan Tompson and Jonathan Yang and Jordi Salvador and Joseph J. Lim and Junhyek Han and Kaiyuan Wang and Kanishka Rao and Karl Pertsch and Karol Hausman and Keegan Go and Keerthana Gopalakrishnan and Ken Goldberg and Kendra Byrne and Kenneth Oslund and Kento Kawaharazuka and Kevin Black and Kevin Lin and Kevin Zhang and Kiana Ehsani and Kiran Lekkala and Kirsty Ellis and Krishan Rana and Krishnan Srinivasan and Kuan Fang and Kunal Pratap Singh and Kuo-Hao Zeng and Kyle Hatch and Kyle Hsu and Laurent Itti and Lawrence Yunliang Chen and Lerrel Pinto and Li Fei-Fei and Liam Tan and Linxi "Jim" Fan and Lionel Ott and Lisa Lee and Luca Weihs and Magnum Chen and Marion Lepert and Marius Memmel and Masayoshi Tomizuka and Masha Itkina and Mateo Guaman Castro and Max Spero and Maximilian Du and Michael Ahn and Michael C. Yip and Mingtong Zhang and Mingyu Ding and Minho Heo and Mohan Kumar Srirama and Mohit Sharma and Moo Jin Kim and Naoaki Kanazawa and Nicklas Hansen and Nicolas Heess and Nikhil J Joshi and Niko Suenderhauf and Ning Liu and Norman Di Palo and Nur Muhammad Mahi Shafiullah and Oier Mees and Oliver Kroemer and Osbert Bastani and Pannag R Sanketi and Patrick "Tree" Miller and Patrick Yin and Paul Wohlhart and Peng Xu and Peter David Fagan and Peter Mitrano and Pierre Sermanet and Pieter Abbeel and Priya Sundaresan and Qiuyu Chen and Quan Vuong and Rafael Rafailov and Ran Tian and Ria Doshi and Roberto Mart{'i}n-Mart{'i}n and Rohan Baijal and Rosario Scalise and Rose Hendrix and Roy Lin and Runjia Qian and Ruohan Zhang and Russell Mendonca and Rutav Shah and Ryan Hoque and Ryan Julian and Samuel Bustamante and Sean Kirmani and Sergey Levine and Shan Lin and Sherry Moore and Shikhar Bahl and Shivin Dass and Shubham Sonawani and Shubham Tulsiani and Shuran Song and Sichun Xu and Siddhant Haldar and Siddharth Karamcheti and Simeon Adebola and Simon Guist and Soroush Nasiriany and Stefan Schaal and Stefan Welker and Stephen Tian and Subramanian Ramamoorthy and Sudeep Dasari and Suneel Belkhale and Sungjae Park and Suraj Nair and Suvir Mirchandani and Takayuki Osa and Tanmay Gupta and Tatsuya Harada and Tatsuya Matsushima and Ted Xiao and Thomas Kollar and Tianhe Yu and Tianli Ding and Todor Davchev and Tony Z. Zhao and Travis Armstrong and Trevor Darrell and Trinity Chung and Vidhi Jain and Vikash Kumar and Vincent Vanhoucke and Wei Zhan and Wenxuan Zhou and Wolfram Burgard and Xi Chen and Xiangyu Chen and Xiaolong Wang and Xinghao Zhu and Xinyang Geng and Xiyuan Liu and Xu Liangwei and Xuanlin Li and Yansong Pang and Yao Lu and Yecheng Jason Ma and Yejin Kim and Yevgen Chebotar and Yifan Zhou and Yifeng Zhu and Yilin Wu and Ying Xu and Yixuan Wang and Yonatan Bisk and Yongqiang Dou and Yoonyoung Cho and Youngwoon Lee and Yuchen Cui and Yue Cao and Yueh-Hua Wu and Yujin Tang and Yuke Zhu and Yunchu Zhang and Yunfan Jiang and Yunshuang Li and Yunzhu Li and Yusuke Iwasawa and Yutaka Matsuo and Zehan Ma and Zhuo Xu and Zichen Jeff Cui and Zichen Zhang and Zipeng Fu and Zipeng Lin},
  booktitle=ICRA,
  year={2024},
}

@inproceedings{liang2024mixtureoftransformerssparsescalablearchitecture,
      title={Mixture-of-Transformers: A Sparse and Scalable Architecture for Multi-Modal Foundation Models}, 
      author={Weixin Liang and Lili Yu and Liang Luo and Srinivasan Iyer and Ning Dong and Chunting Zhou and Gargi Ghosh and Mike Lewis and Wen-tau Yih and Luke Zettlemoyer and Xi Victoria Lin},
      year={2024},
      booktitle=CPAL,
}

@inproceedings{liu2024rdt,
    title={RDT-1B: a Diffusion Foundation Model for Bimanual Manipulation},
    author={Liu, Songming and Wu, Lingxuan and Li, Bangguo and Tan, Hengkai and Chen, Huayu and Wang, Zhengyi and Xu, Ke and Su, Hang and Zhu, Jun},
    booktitle=ICLR,
    year={2024}
}

@inproceedings{rt12022arxiv,
    title={RT-1: Robotics Transformer for Real-World Control at Scale},
    author={Anthony	Brohan and  Noah Brown and  Justice Carbajal and  Yevgen Chebotar and  Joseph Dabis and  Chelsea Finn and  Keerthana Gopalakrishnan and  Karol Hausman and  Alex Herzog and  Jasmine Hsu and  Julian Ibarz and  Brian Ichter and  Alex Irpan and  Tomas Jackson and  Sally Jesmonth and  Nikhil Joshi and  Ryan Julian and  Dmitry Kalashnikov and  Yuheng Kuang and  Isabel Leal and  Kuang-Huei Lee and  Sergey Levine and  Yao Lu and  Utsav Malla and  Deeksha Manjunath and  Igor Mordatch and  Ofir Nachum and  Carolina Parada and  Jodilyn Peralta and  Emily Perez and  Karl Pertsch and  Jornell Quiambao and  Kanishka Rao and  Michael Ryoo and  Grecia Salazar and  Pannag Sanketi and  Kevin Sayed and  Jaspiar Singh and  Sumedh Sontakke and  Austin Stone and  Clayton Tan and  Huong Tran and  Vincent Vanhoucke and Steve Vega and  Quan Vuong and  Fei Xia and  Ted Xiao and  Peng Xu and  Sichun Xu and  Tianhe Yu and  Brianna Zitkovich},
    booktitle=RSS,
    year={2022}
}

@inproceedings{rt22023arxiv,
  title={Rt-2: Vision-language-action models transfer web knowledge to robotic control},
  author={Zitkovich, Brianna and Yu, Tianhe and Xu, Sichun and Xu, Peng and Xiao, Ted and Xia, Fei and Wu, Jialin and Wohlhart, Paul and Welker, Stefan and Wahid, Ayzaan and others},
  booktitle=CoRL,
  pages={2165--2183},
  year={2023},
  organization={PMLR}
}

@inproceedings{octo_2023,
    title={Octo: An Open-Source Generalist Robot Policy},
    author = {{Octo Model Team} and Dibya Ghosh and Homer Walke and Karl Pertsch and Kevin Black and Oier Mees and Sudeep Dasari and Joey Hejna and Charles Xu and Jianlan Luo and Tobias Kreiman and {You Liang} Tan and Lawrence Yunliang Chen and Pannag Sanketi and Quan Vuong and Ted Xiao and Dorsa Sadigh and Chelsea Finn and Sergey Levine},
    booktitle = RSS,
    address  = {Delft, Netherlands},
    year = {2024},
}

@inproceedings{doshi2024scalingcrossembodiedlearningpolicy,
  title={Scaling Cross-Embodied Learning: One Policy for Manipulation, Navigation, Locomotion and Aviation},
  author={Doshi, Ria and Walke, Homer Rich and Mees, Oier and Dasari, Sudeep and Levine, Sergey},
  booktitle=CoRL,
  year = {2024}
}

@inproceedings{walke2023bridgedata,
    title={BridgeData V2: A Dataset for Robot Learning at Scale},
    author={Walke, Homer and Black, Kevin and Lee, Abraham and Kim, Moo Jin and Du, Max and Zheng, Chongyi and Zhao, Tony and Hansen-Estruch, Philippe and Vuong, Quan and He, Andre and Myers, Vivek and Fang, Kuan and Finn, Chelsea and Levine, Sergey},
    booktitle=CoRL,
    year={2023}
}

@inproceedings{rosete2022tacorl,
    author = {Erick Rosete-Beas and Oier Mees and Gabriel Kalweit and Joschka Boedecker and Wolfram Burgard},
    title = {Latent Plans for Task Agnostic Offline Reinforcement Learning},
    booktitle = CoRL,
    year = {2022}
}

@article{zhu2022viola,
  title={VIOLA: Imitation Learning for Vision-Based Manipulation with Object Proposal Priors},
  author={Zhu, Yifeng and Joshi, Abhishek and Stone, Peter and Zhu, Yuke},
  journal=CoRL,
  year={2022}
}

@misc{dass2023jacoplay,
  author = {Dass, Shivin and Yapeter, Jullian and Zhang, Jesse and Zhang, Jiahui
            and Pertsch, Karl and Nikolaidis, Stefanos and Lim, Joseph J.},
  title = {CLVR Jaco Play Dataset},
  url = {https://github.com/clvrai/clvr_jaco_play_dataset},
  version = {1.0.0},
  year = {2023}
}

@misc{BerkeleyUR5Website,
  title = {Berkeley {UR5} Demonstration Dataset},
  author = {Lawrence Yunliang Chen and Simeon Adebola and Ken Goldberg},
  howpublished = {\url{https://sites.google.com/view/berkeley-ur5/home}},
  year = {2023}
}

@inproceedings{liu2023libero,
  title={Libero: Benchmarking knowledge transfer for lifelong robot learning},
  author={Liu, Bo and Zhu, Yifeng and Gao, Chongkai and Feng, Yihao and Liu, Qiang and Zhu, Yuke and Stone, Peter},
  booktitle=NeurIPS,
  volume={36},
  pages={44776--44791},
  year={2023}
}

@misc{nvidia2025gr00tn1openfoundation,
      title={GR00T N1: An Open Foundation Model for Generalist Humanoid Robots}, 
      author={NVIDIA and Johan Bjorck and Fernando Castañeda and Nikita Cherniadev and Xingye Da and Runyu Ding and Linxi "Jim" Fan and Yu Fang and Dieter Fox and Fengyuan Hu and Spencer Huang and Joel Jang and Zhenyu Jiang and Jan Kautz and Kaushil Kundalia and Lawrence Lao and Zhiqi Li and Zongyu Lin and Kevin Lin and Guilin Liu and Edith Llontop and Loic Magne and Ajay Mandlekar and Avnish Narayan and Soroush Nasiriany and Scott Reed and You Liang Tan and Guanzhi Wang and Zu Wang and Jing Wang and Qi Wang and Jiannan Xiang and Yuqi Xie and Yinzhen Xu and Zhenjia Xu and Seonghyeon Ye and Zhiding Yu and Ao Zhang and Hao Zhang and Yizhou Zhao and Ruijie Zheng and Yuke Zhu},
      year={2025},
      eprint={2503.14734},
      archivePrefix={arXiv},
      primaryClass={cs.RO},
      url={https://arxiv.org/abs/2503.14734}, 
}

@inproceedings{10.5555/3692070.3694018,
    author = {Tang, Anke and Shen, Li and Luo, Yong and Yin, Nan and Zhang, Lefei and Tao, Dacheng},
    title = {Merging multi-task models via weight-ensembling mixture of experts},
    year = {2024},
    booktitle = ICML,
}

@inproceedings{kim2025fine,
  title={Fine-Tuning Vision-Language-Action Models: Optimizing Speed and Success},
  author={Kim, Moo Jin and Finn, Chelsea and Liang, Percy},
  booktitle=RSS,
  year={2025}
}

@article{li2025eagle2buildingposttraining,
      title={Eagle 2: Building Post-Training Data Strategies from Scratch for Frontier Vision-Language Models}, 
      author={Zhiqi Li and Guo Chen and Shilong Liu and Shihao Wang and Vibashan VS and Yishen Ji and Shiyi Lan and Hao Zhang and Yilin Zhao and Subhashree Radhakrishnan and Nadine Chang and Karan Sapra and Amala Sanjay Deshmukh and Tuomas Rintamaki and Matthieu Le and Ilia Karmanov and Lukas Voegtle and Philipp Fischer and De-An Huang and Timo Roman and Tong Lu and Jose M. Alvarez and Bryan Catanzaro and Jan Kautz and Andrew Tao and Guilin Liu and Zhiding Yu},
      year={2025},
      journal={CoRR}
}

@article{li2024cogactfoundationalvisionlanguageactionmodel,
      title={CogACT: A Foundational Vision-Language-Action Model for Synergizing Cognition and Action in Robotic Manipulation}, 
      author={Qixiu Li and Yaobo Liang and Zeyu Wang and Lin Luo and Xi Chen and Mozheng Liao and Fangyun Wei and Yu Deng and Sicheng Xu and Yizhong Zhang and Xiaofan Wang and Bei Liu and Jianlong Fu and Jianmin Bao and Dong Chen and Yuanchun Shi and Jiaolong Yang and Baining Guo},
      year={2024},
      journal={CoRR}
}

@inproceedings{Peebles2022ScalableDM,
  title={Scalable Diffusion Models with Transformers},
  author={William S. Peebles and Saining Xie},
  booktitle=ICCV,
  year={2023},
  pages={4172-4182}
}

@inproceedings{8953486,
  author={Zhou, Yi and Barnes, Connelly and Lu, Jingwan and Yang, Jimei and Li, Hao},
  booktitle=CVPR, 
  title={On the Continuity of Rotation Representations in Neural Networks}, 
  year={2019},
  volume={},
  number={},
  pages={5738-5746},
  doi={10.1109/CVPR.2019.00589}}

@inproceedings{pertsch2025fastefficientactiontokenization,
      title={FAST: Efficient Action Tokenization for Vision-Language-Action Models}, 
      author={Karl Pertsch and Kyle Stachowicz and Brian Ichter and Danny Driess and Suraj Nair and Quan Vuong and Oier Mees and Chelsea Finn and Sergey Levine},
      year={2025},
      booktitle=RSS
}

@article{1672377,
  author={Ahmed, N. and Natarajan, T. and Rao, K.R.},
  journal=TC, 
  title={Discrete Cosine Transform}, 
  year={1974},
  volume={C-23},
  number={1},
  pages={90-93},
  doi={10.1109/T-C.1974.223784}
}

@article{tunyasuvunakool2020,
         title = {dm\_control: Software and tasks for continuous control},
         journal = {Software Impacts},
         volume = {6},
         pages = {100022},
         year = {2020},
         issn = {2665-9638},
         doi = {https://doi.org/10.1016/j.simpa.2020.100022},
         url = {https://www.sciencedirect.com/science/article/pii/S2665963820300099},
         author = {Saran Tunyasuvunakool and Alistair Muldal and Yotam Doron and
                   Siqi Liu and Steven Bohez and Josh Merel and Tom Erez and
                   Timothy Lillicrap and Nicolas Heess and Yuval Tassa},
}

@misc{aloha2team2024aloha2enhancedlowcost,
      title={ALOHA 2: An Enhanced Low-Cost Hardware for Bimanual Teleoperation}, 
      author={ALOHA 2 Team and Jorge Aldaco and Travis Armstrong and Robert Baruch and Jeff Bingham and Sanky Chan and Kenneth Draper and Debidatta Dwibedi and Chelsea Finn and Pete Florence and Spencer Goodrich and Wayne Gramlich and Torr Hage and Alexander Herzog and Jonathan Hoech and Thinh Nguyen and Ian Storz and Baruch Tabanpour and Leila Takayama and Jonathan Tompson and Ayzaan Wahid and Ted Wahrburg and Sichun Xu and Sergey Yaroshenko and Kevin Zakka and Tony Z. Zhao},
      year={2024},
      eprint={2405.02292},
      archivePrefix={arXiv},
      primaryClass={cs.RO},
      url={https://arxiv.org/abs/2405.02292}, 
}

@inproceedings{DBLP:journals/corr/abs-2204-11918,
  title={Google scanned objects: A high-quality dataset of 3d scanned household items},
  author={Downs, Laura and Francis, Anthony and Koenig, Nate and Kinman, Brandon and Hickman, Ryan and Reymann, Krista and McHugh, Thomas B and Vanhoucke, Vincent},
  booktitle=ICRA,
  pages={2553--2560},
  year={2022},
  organization={IEEE}
}

@inproceedings{wu2024gellogenerallowcostintuitive,
  title={Gello: A general, low-cost, and intuitive teleoperation framework for robot manipulators},
  author={Wu, Philipp and Shentu, Yide and Yi, Zhongke and Lin, Xingyu and Abbeel, Pieter},
  booktitle=IROS,
  pages={12156--12163},
  year={2024},
  organization={IEEE}
}

@inproceedings{mu2025robotwindualarmrobotbenchmark,
  author = {Mu, Yao and Chen, Tianxing and Chen, Zanxin and Peng, Shijia and Lan, Zhiqian and Gao, Zeyu and Liang, Zhixuan and Yu, Qiaojun and Zou, Yude and Xu, Mingkun and Lin, Lunkai and Xie, Zhiqiang and Ding, Mingyu and Luo, Ping},
  title = {RoboTwin: Dual-Arm Robot Benchmark with Generative Digital Twins},
  booktitle = CVPR,
  month = {June},
  year = {2025},
  pages = {27649-27660}
}

@inproceedings{chi2024universalmanipulationinterfaceinthewild,
  title={Universal Manipulation Interface: In-The-Wild Robot Teaching Without In-The-Wild Robots},
  author={Chi, Cheng and Xu, Zhenjia and Pan, Chuer and Cousineau, Eric and Burchfiel, Benjamin and Feng, Siyuan and Tedrake, Russ and Song, Shuran},
  booktitle=RSS,
  year={2024}
}

@inproceedings{xie2020deepimitationlearningbimanual,
  title={Deep imitation learning for bimanual robotic manipulation},
  author={Xie, Fan and Chowdhury, Alexander and De Paolis Kaluza, M and Zhao, Linfeng and Wong, Lawson and Yu, Rose},
  booktitle=NeurIPS,
  volume={33},
  pages={2327--2337},
  year={2020}
}

@inproceedings{liu2023visualinstructiontuning,
  title={Visual instruction tuning},
  author={Liu, Haotian and Li, Chunyuan and Wu, Qingyang and Lee, Yong Jae},
  booktitle=NeurIPS,
  volume={36},
  pages={34892--34916},
  year={2023}
}

@inproceedings{li2024visionlanguagefoundationmodelseffective,
  title={Vision-Language Foundation Models as Effective Robot Imitators},
  author={Li, Xinghang and Liu, Minghuan and Zhang, Hanbo and Yu, Cunjun and Xu, Jie and Wu, Hongtao and Cheang, Chilam and Jing, Ya and Zhang, Weinan and Liu, Huaping and others},
  booktitle=ICLR,
  year={2024}
}

@misc{cadene2024lerobot,
    author = {Cadene, Remi and Alibert, Simon and Soare, Alexander and Gallouedec, Quentin and Zouitine, Adil and Wolf, Thomas},
    title = {LeRobot: State-of-the-art Machine Learning for Real-World Robotics in Pytorch},
    howpublished = "\url{https://github.com/huggingface/lerobot}",
    year = {2024}
}

@inproceedings{lee2020learning,
  title={Learning to Coordinate Manipulation Skills via Skill Behavior Diversification},
  author={Youngwoon Lee and Jingyun Yang and Joseph J. Lim},
  booktitle=ICLR,
  year={2020},
}

@inproceedings{stavridis2018bimanual,
  title={Bimanual assembly of two parts with relative motion generation and task related optimization},
  author={Stavridis, Sotiris and Doulgeri, Zoe},
  booktitle=IROS,
  pages={7131--7136},
  year={2018},
  organization={IEEE}
}

@inproceedings{bersch2011bimanual,
  title={Bimanual robotic cloth manipulation for laundry folding},
  author={Bersch, Christian and Pitzer, Benjamin and Kammel, S{\"o}ren},
  booktitle=IROS,
  pages={1413--1419},
  year={2011},
  organization={IEEE}
}

@inproceedings{avigal2022speedfolding,
  title={Speedfolding: Learning efficient bimanual folding of garments},
  author={Avigal, Yahav and Berscheid, Lars and Asfour, Tamim and Kr{\"o}ger, Torsten and Goldberg, Ken},
  booktitle=IROS,
  pages={1--8},
  year={2022},
  organization={IEEE}
}

@article{Yadav_Nagar_Shah_2024,
  title={Learning vision-based robotic manipulation tasks sequentially in offline reinforcement learning settings},
  volume={42},
  number={6},
  journal={Robotica},
  author={Yadav, Sudhir Pratap and Nagar, Rajendra and Shah, Suril V.},
  year={2024},
  pages={1715–1730}
}

@inproceedings{belkhale2023hydra,
 title={HYDRA: Hybrid Robot Actions for Imitation Learning},
 author={Belkhale, Suneel and Cui, Yuchen and Sadigh, Dorsa},
 booktitle=CoRL,
 year={2023}
}

@article{9695333,
  author={Zhu, Yifeng and Stone, Peter and Zhu, Yuke},
  journal=RAL, 
  title={Bottom-Up Skill Discovery From Unsegmented Demonstrations for Long-Horizon Robot Manipulation}, 
  year={2022},
  volume={7},
  number={2},
  pages={4126-4133},
  doi={10.1109/LRA.2022.3146589}
}

@inproceedings{heo2023furniturebench,
  title={FurnitureBench: Reproducible Real-World Benchmark for Long-Horizon Complex Manipulation},
  author={Minho Heo and Youngwoon Lee and Doohyun Lee and Joseph J. Lim},
  booktitle=RSS,
  year={2023}
}

@inproceedings{shah2023mutex,
	title        = {MUTEX: Learning Unified Policies from Multimodal Task Specifications},
	author       = {Rutav Shah and Roberto Mart{\'\i}n-Mart{\'\i}n and Yuke Zhu},
	year         = 2023,
	booktitle    = CoRL,
	url          = {https://openreview.net/forum?id=PwqiqaaEzJ}
}

@article{doi:10.1177/02783649241276017,
  author={Jianlan Luo and Charles Xu and Fangchen Liu and Liam Tan and Zipeng Lin and Jeffrey Wu and Pieter Abbeel and Sergey Levine},
  title={FMB: A functional manipulation benchmark for generalizable robotic learning},
  journal=IJRR,
  volume={44},
  number={4},
  pages={592-606},
  year={2025}
}

@misc{shafiullah2023bringing,
  title={On bringing robots home},
  author={Shafiullah, Nur Muhammad Mahi and Rai, Anant and Etukuru, Haritheja and Liu, Yiqian and Misra, Ishan and Chintala, Soumith and Pinto, Lerrel},
  journal={arXiv preprint arXiv:2311.16098},
  year={2023}
}

@inproceedings{jang2021bc,   
    title={{BC}-Z: Zero-Shot Task Generalization with Robotic Imitation Learning},
    author={Eric Jang and Alex Irpan and Mohi Khansari and Daniel Kappler and Frederik Ebert and Corey Lynch and Sergey Levine and Chelsea Finn},
    booktitle=CoRL,
    year={2021},
}

@inproceedings{bahl2023affordances,
  title={Affordances from Human Videos as a Versatile Representation for Robotics},
  author={Bahl, Shikhar and Mendonca, Russell and Chen, Lili and Jain, Unnat and Pathak, Deepak},
  booktitle=CVPR,
  year={2023}
}

@article{mendonca2023structured,
  title={Structured World Models from Human Videos},
  author={Mendonca, Russell and Bahl, Shikhar and Pathak, Deepak},
  journal=CoRL,
  year={2023}
}

@misc{zhu2023fanuc, 
  title={Fanuc Manipulation: A Dataset for Learning-based Manipulation with FANUC Mate 200iD Robot}, 
  author={Zhu, Xinghao and Tian, Ran and Xu, Chenfeng and Huo, Mingxiao and Zhan, Wei and Tomizuka, Masayoshi and Ding, Mingyu}, 
  howpublished={\url{https://sites.google.com/berkeley.edu/fanuc-manipulation}}, 
  year={2023} 
}

@inproceedings{fu2024mobile,
  author    = {Fu, Zipeng and Zhao, Tony Z. and Finn, Chelsea},
  title     = {Mobile ALOHA: Learning Bimanual Mobile Manipulation with Low-Cost Whole-Body Teleoperation},
  booktitle = CoRL,
  year      = {2024},
}

@inproceedings{cui2022play,
    title   = {From Play to Policy: Conditional Behavior Generation from Uncurated Robot Data},
    author  = {Cui, Zichen Jeff and Wang, Yibin and Shafiullah, Nur Muhammad Mahi and Pinto, Lerrel},
    booktitle = ICLR,
    year    = {2023}
}

@inproceedings{nasiriany2022sailor,
  title={Learning and Retrieval from Prior Data for Skill-based Imitation Learning},
  author={Soroush Nasiriany and Tian Gao and Ajay Mandlekar and Yuke Zhu},
  booktitle=CoRL,
  year={2022}
}

@inproceedings{liu2022robot,
    title = {Robot Learning on the Job: Human-in-the-Loop Autonomy and Learning During Deployment},
    author = {Huihan Liu and Soroush Nasiriany and Lance Zhang and Zhiyao Bao and Yuke Zhu},
    booktitle = RSS,
    year = {2023}
}

@inproceedings{vogel_edan_2020,
        title = {EDAN - an EMG-Controlled Daily Assistant to Help People with Physical Disabilities},
        language = {en},
        booktitle = IROS,
        author = {Vogel, Jörn and Hagengruber, Annette and Iskandar, Maged and Quere, Gabriel and Leipscher, Ulrike and Bustamante, Samuel and Dietrich, Alexander and Hoeppner, Hannes and Leidner, Daniel and Albu-Schäffer, Alin},
        year = {2020}
}

@inproceedings{quere_shared_2020,
        address = {Paris, France},
        title = {Shared {Control} {Templates} for {Assistive} {Robotics}},
        language = {en},
        booktitle = ICRA,
        author = {Quere, Gabriel and Hagengruber, Annette and Iskandar, Maged and Bustamante, Samuel and Leidner, Daniel and Stulp, Freek and Vogel, Joern},
        year = {2020},
}

@article{chen2025expandingperformanceboundariesopensource,
      title={Expanding Performance Boundaries of Open-Source Multimodal Models with Model, Data, and Test-Time Scaling}, 
      author={Zhe Chen and Weiyun Wang and Yue Cao and Yangzhou Liu and Zhangwei Gao and Erfei Cui and Jinguo Zhu and Shenglong Ye and Hao Tian and Zhaoyang Liu and Lixin Gu and Xuehui Wang and Qingyun Li and Yimin Ren and Zixuan Chen and Jiapeng Luo and Jiahao Wang and Tan Jiang and Bo Wang and Conghui He and Botian Shi and Xingcheng Zhang and Han Lv and Yi Wang and Wenqi Shao and Pei Chu and Zhongying Tu and Tong He and Zhiyong Wu and Huipeng Deng and Jiaye Ge and Kai Chen and Kaipeng Zhang and Limin Wang and Min Dou and Lewei Lu and Xizhou Zhu and Tong Lu and Dahua Lin and Yu Qiao and Jifeng Dai and Wenhai Wang},
      year={2025},
      journal={CoRR}
}

@article{wang2024qwen2vlenhancingvisionlanguagemodels,
      title={Qwen2-VL: Enhancing Vision-Language Model's Perception of the World at Any Resolution}, 
      author={Peng Wang and Shuai Bai and Sinan Tan and Shijie Wang and Zhihao Fan and Jinze Bai and Keqin Chen and Xuejing Liu and Jialin Wang and Wenbin Ge and Yang Fan and Kai Dang and Mengfei Du and Xuancheng Ren and Rui Men and Dayiheng Liu and Chang Zhou and Jingren Zhou and Junyang Lin},
      year={2024},
      journal={CoRR}
}

@inproceedings{shazeer2017outrageously,
  title={Outrageously Large Neural Networks: The Sparsely-Gated Mixture-of-Experts Layer},
  author={Shazeer, Noam and Mirhoseini, Azalia and Maziarz, Krzysztof and Davis, Andy and Le, Quoc and Hinton, Geoffrey and Dean, Jeff},
  booktitle=ICLR,
  year={2017},
  url={https://openreview.net/forum?id=B1ckMDqlg},
}

@article{chen2025robotwin,
  title={RoboTwin 2.0: A Scalable Data Generator and Benchmark with Strong Domain Randomization for Robust Bimanual Robotic Manipulation},
  author={Chen, Tianxing and Chen, Zanxin and Chen, Baijun and Cai, Zijian and Liu, Yibin and Liang, Qiwei and Li, Zixuan and Lin, Xianliang and Ge, Yiheng and Gu, Zhenyu and others},
  journal={arXiv preprint arXiv:2506.18088},
  year={2025}
}

@inproceedings{anubis2025,
  title={ANUBIS: A Compact, Low-Cost, Compliant Humanoid Mobile Manipulation Robot},
  author={Kang, Taewoong and Kim, Joonyoung and Nasrat, Shady and Song, Dongwoon and Ahn, Gijae and Jo, Minseong and Lee, Seonil and Yi, Seung-Joon},
  booktitle=Humanoids,
  year={2025},
  url={https://ras.papercept.net/conferences/scripts/rtf/ICHR25_ContentListWeb_2.html},
}

@article {Sadato9667,
	author = {Sadato, Norihiro and Yonekura, Yoshiharu and Waki, Atsuo and Yamada, Hiroki and Ishii, Yasushi},
	title = {Role of the Supplementary Motor Area and the Right Premotor Cortex in the Coordination of Bimanual Finger Movements},
	volume = {17},
	number = {24},
	pages = {9667--9674},
	year = {1997},
	doi = {10.1523/JNEUROSCI.17-24-09667.1997},
	publisher = {Society for Neuroscience},
	issn = {0270-6474},
	URL = {https://www.jneurosci.org/content/17/24/9667},
	eprint = {https://www.jneurosci.org/content/17/24/9667.full.pdf},
	journal = {Journal of Neuroscience}
}

@article{Swinnen2002,
  author  = {Swinnen, Stephan P.},
  title   = {Intermanual coordination: From behavioural principles to neural-network interactions},
  journal = {Nature Reviews Neuroscience},
  year    = {2002},
  volume  = {3},
  number  = {5},
  pages   = {348--359},
  doi     = {10.1038/nrn807}
}

@inproceedings{
    black2025pi,
    title={\${\textbackslash}pi\_\{0.5\}\$: a Vision-Language-Action Model with Open-World Generalization},
    author={Kevin Black and Noah Brown and James Darpinian and Karan Dhabalia and Danny Driess and Adnan Esmail and Michael Robert Equi and Chelsea Finn and Niccolo Fusai and Manuel Y. Galliker and Dibya Ghosh and Lachy Groom and Karol Hausman and brian ichter and Szymon Jakubczak and Tim Jones and Liyiming Ke and Devin LeBlanc and Sergey Levine and Adrian Li-Bell and Mohith Mothukuri and Suraj Nair and Karl Pertsch and Allen Z. Ren and Lucy Xiaoyang Shi and Laura Smith and Jost Tobias Springenberg and Kyle Stachowicz and James Tanner and Quan Vuong and Homer Walke and Anna Walling and Haohuan Wang and Lili Yu and Ury Zhilinsky},
    booktitle=CoRL,
    year={2025},
    url={https://openreview.net/forum?id=vlhoswksBO}
}

@inproceedings{11127809,
  author={Jiang, Zhenyu and Xie, Yuqi and Lin, Kevin and Xu, Zhenjia and Wan, Weikang and Mandlekar, Ajay and Fan, Linxi Jim and Zhu, Yuke},
  booktitle=ICRA, 
  title={DexMimicGen: Automated Data Generation for Bimanual Dexterous Manipulation via Imitation Learning}, 
  year={2025},
  volume={},
  number={},
  pages={16923-16930},
  doi={10.1109/ICRA55743.2025.11127809}}

@misc{aloha_sim_github,
  author       = {{Google DeepMind}},
  title        = {Aloha-Sim},
  howpublished = {\url{https://github.com/google-deepmind/aloha_sim}},
  year         = {2025},
  note         = {Accessed: 2025-10-24}
}

@inproceedings{lu2025anybimanualtransferringunimanualpolicy,
      title={AnyBimanual: Transferring Unimanual Policy for General Bimanual Manipulation}, 
      author={Guanxing Lu and Tengbo Yu and Haoyuan Deng and Season Si Chen and Yansong Tang and Ziwei Wang},
      year={2025},
      booktitle=ICCV,
}

@misc{kobayashi2025bivlabilateralcontrolbasedimitation,
      title={Bi-VLA: Bilateral Control-Based Imitation Learning via Vision-Language Fusion for Action Generation}, 
      author={Masato Kobayashi and Thanpimon Buamanee},
      year={2025},
      eprint={2509.18865},
      archivePrefix={arXiv},
      primaryClass={cs.RO},
      url={https://arxiv.org/abs/2509.18865}, 
}

@inproceedings{10831380,
  author={Gbagbe, Koffivi Fidèle and Cabrera, Miguel Altamirano and Alabbas, Ali and Alyunes, Oussama and Lykov, Artem and Tsetserukou, Dzmitry},
  booktitle={IEEE International Conference on Systems, Man, and Cybernetics (SMC)}, 
  title={Bi-VLA: Vision-Language-Action Model-Based System for Bimanual Robotic Dexterous Manipulations}, 
  year={2024},
  volume={},
  number={},
  pages={2864-2869},
  doi={10.1109/SMC54092.2024.10831380}
}

@article{DBLP:journals/corr/abs-2407-00278,
  publtype={informal},
  author={Markus Grotz and Mohit Shridhar and Tamim Asfour and Dieter Fox},
  title={PerAct2: Benchmarking and Learning for Robotic Bimanual Manipulation Tasks},
  year={2024},
  cdate={1704067200000},
  journal={CoRR},
  volume={abs/2407.00278},
  url={https://doi.org/10.48550/arXiv.2407.00278}
}

@inproceedings{shridhar2022peract,
  title     = {Perceiver-Actor: A Multi-Task Transformer for Robotic Manipulation}, 
  author    = {Shridhar, Mohit and Manuelli, Lucas and Fox, Dieter},
  booktitle = CoRL,
  year      = {2022},
}
